\documentclass[10pt]{article} 
\PassOptionsToPackage{dvipsnames}{xcolor}
\usepackage[accepted]{tmlr}

\usepackage{amsmath,amsfonts,bm}

\def\eqref#1{equation~\ref{#1}}

\def\1{\bm{1}}

\DeclareMathAlphabet{\mathsfit}{\encodingdefault}{\sfdefault}{m}{sl}
\SetMathAlphabet{\mathsfit}{bold}{\encodingdefault}{\sfdefault}{bx}{n}

\usepackage[normalem]{ulem}
\usepackage[utf8]{inputenc}
\usepackage[T1]{fontenc}
\usepackage{hyperref}
\usepackage{url}
\usepackage{booktabs}
\usepackage{amsfonts}
\usepackage{nicefrac}
\usepackage{microtype}
\usepackage{graphicx}
\usepackage{grffile}
\usepackage{subcaption}
\usepackage{amsmath}
\usepackage{amssymb}
\usepackage{mathtools}
\usepackage{amsthm}
\usepackage{listings}
\usepackage{framed}
\usepackage{booktabs,multirow}
\usepackage[capitalize,noabbrev]{cleveref}
\usepackage{tabularx}
\usepackage{placeins}

\let\origsubsection\subsection
\renewcommand{\subsection}{\FloatBarrier\origsubsection}

\newcolumntype{C}{>{\centering\arraybackslash}X}

\newif\ifrevisionmarkup
\revisionmarkupfalse
\ifrevisionmarkup
  \newcommand{\add}[1]{{\color{blue}#1}}
  \newcommand{\del}[1]{{\color{BrickRed}\sout{#1}}}
\else
  \newcommand{\add}[1]{#1}
  \newcommand{\del}[1]{}
\fi
\newenvironment{addblock}{\ifrevisionmarkup\color{blue}\fi}{}
\newenvironment{delblock}{\ifrevisionmarkup\color{BrickRed}\fi}{}
\newcommand{\tablemarkup}{\ifrevisionmarkup\color{blue}\fi}

\theoremstyle{plain}

\theoremstyle{definition}

\theoremstyle{remark}

\title{\texorpdfstring{%
Demasking Fine-Tuning for Knowledge Injection in \\ Autoregressive and Diffusion LLMs}%
{Demasking Fine-Tuning for Knowledge Injection in Autoregressive and Diffusion LLMs}}

\author{\name Xu Pan\thanks{Equal contribution.} \email xupan@fas.harvard.edu \\
      \addr Harvard University
      \AND
      \name Ely Hahami\footnotemark[1] \email elyhahami@college.harvard.edu \\
      \addr Harvard University
      \AND
      \name Jingxuan Fan\footnotemark[1] \email jfan@g.harvard.edu \\
      \addr Harvard University
      \AND
      \name Ziqian Xie 
      \email ziqian.xie@yale.edu
      \\
      \addr Yale University \\
      University of Texas Health Science Center at Houston
      \AND
      \name Haim Sompolinsky
      \email hsompolinsky@mcb.harvard.edu
      \\
      \addr Harvard University \\
      Edmond and Lily Safra Center for Brain Sciences, Hebrew University}

\def\month{11}
\def\year{2026}
\def\openreview{\url{https://openreview.net/forum?id=5R85AMY7tX}}

\begin{document}

\maketitle

\begin{abstract}
Large language models (LLMs) are often used in environments where facts evolve, yet factual knowledge updates via fine-tuning on unstructured plain text often suffer from 1) reliance on paraphrasing augmentation, without which training does not generalize to downstream QA tasks, and 2) the reversal curse. Recent evidence of the sample efficiency of masked diffusion large language models (dLLMs) and the bidirectional information flow in their demasking training objective suggests that the demasking objective may facilitate knowledge learning from raw knowledge text. We test this hypothesis in controlled knowledge fine-tuning experiments on dLLMs and autoregressive LLMs (arLLMs) and find that while arLLMs rely on paraphrase augmentation to generalize knowledge text into question-answering (QA) capability, dLLMs do not require paraphrases to achieve high QA accuracy. We find that giving arLLMs a demasking objective, where the LLM must reconstruct the original text from a masked copy given in context, improves the efficacy of knowledge injection, removing the need for paraphrases and mitigating the reversal curse, recovering most of the gap between arLLMs and dLLMs. We also provide preliminary evidence of the broader applicability of using the demasking objective in arLLM fine-tuning on a large-scale knowledge-intensive dataset. \del{The demasking objective also improves SFT on math tasks, suggesting broad utility beyond factual knowledge injection.}
\end{abstract}

\begin{figure}[t]
  \centering
  \includegraphics[width=0.75\textwidth]{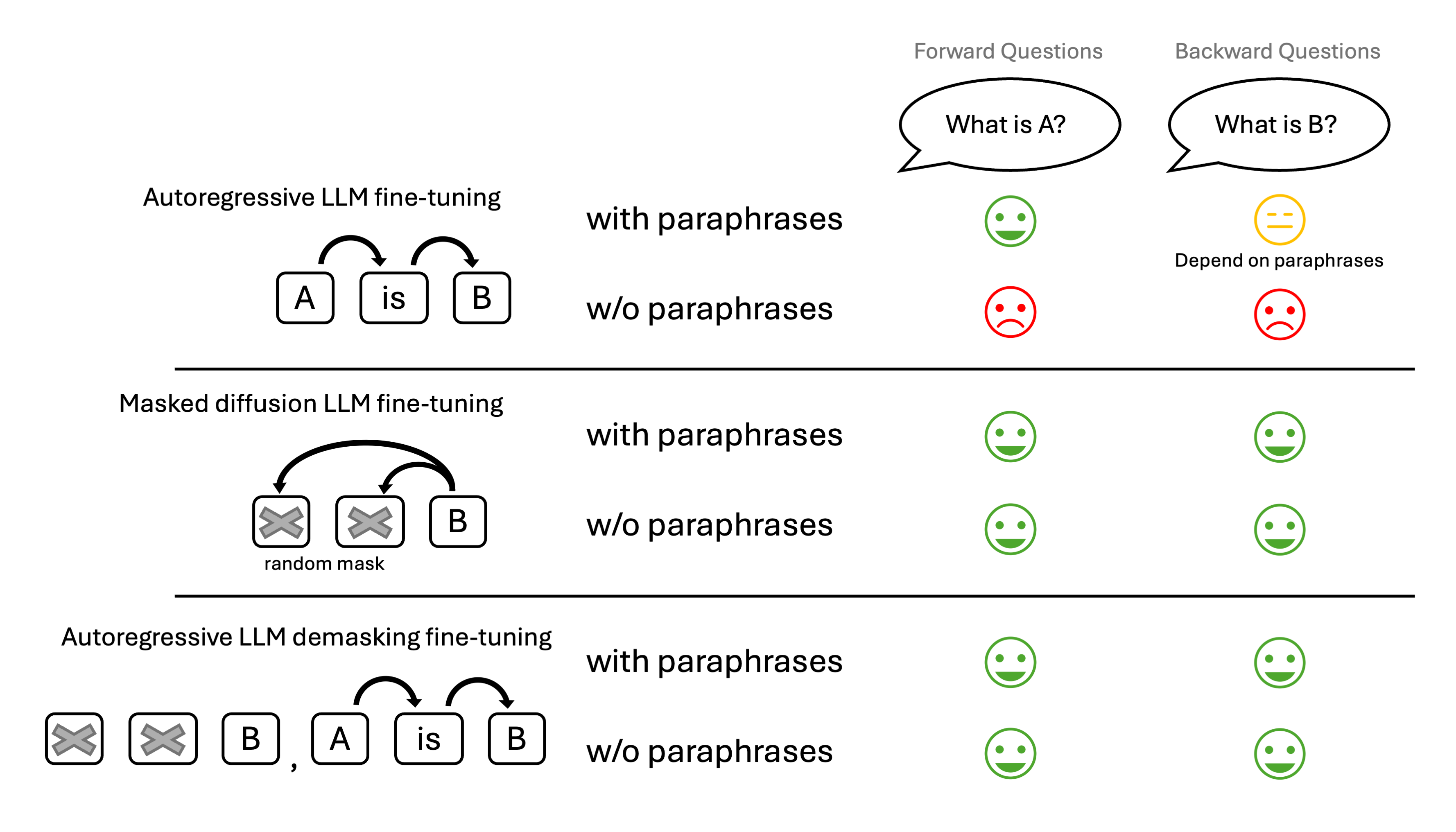}
  \caption{A schematic summary of the results. First row: autoregressive LLM requires paraphrases to generalize knowledge from the fine-tuning text to QA tasks, and suffers from reversal curse (i.e.\ fail to answer backward questions). Second row: masked diffusion LLM can easily generalize fine-tuning text to QA tasks in both forward and backward styles. Third row: demasking fine-tuning of an arLLM recovers most of the fine-tuning gap between autoregressive LLMs and masked diffusion LLMs.}
  \label{fig:diagram}
\end{figure}

\section{Introduction}

Large language models (LLMs) are often deployed in settings where facts evolve: news breaks, policies change, and organizations maintain internal knowledge that is continuously updated. \add{To keep LLM memory up to date, in current practice the mature solution is retrieval-augmented generation (RAG): with relevant documents retrieved into context, models use new knowledge almost perfectly. Despite its practical success, RAG-based memory has intrinsic limitations: a bounded context window and the computational cost of long contexts \citep{liu2023lost}; the growing size of the external storage and the cost of managing its internal consistency; known theoretical bottlenecks of embedding-based retrieval \citep{weller2025theoretical}; and the difficulty of expressing implicit, non-semantic knowledge. Ideally, a lifelong learning system should integrate new knowledge with old knowledge at the level of weights and internal representations, gradually updating its internal view of the world, with the growing knowledge participating in every computation rather than waiting to be retrieved. A naive but natural idea of injecting knowledge into an LLM's weights is to fine-tune the model on newly available documents and expect the model to learn the knowledge and use it flexibly (e.g., answering related questions in varying formats). However, a growing body of work shows that fine-tuning on knowledge documents often struggles to generalize to question-answering tasks, even when the fine-tuning loss decreases \citep{ovadia2023fine, mecklenburg2024injecting, gekhman2024does, soudani2024fine, zhao2025style, lampinen2025generalization, zhu24b}. This limits the practicality of weight updating as a long-term memory mechanism in LLMs.} 

There are at least two empirical obstacles that make knowledge injection via fine-tuning in autoregressive LLMs (arLLMs) difficult to generalize to downstream QA tasks. (1) Reliance on paraphrasing augmentation: after auto-regressive fine-tuning on raw documents, arLLMs often fail to generalize learned knowledge to question-answering behavior, unless the training data is augmented with paraphrases \citep{ovadia2023fine, mecklenburg2024injecting}. Paraphrasing augmentation is expensive and sometimes impractical, as high-quality paraphrases typically require additional LLM calls, careful filtering, and substantial engineering to avoid distribution shift. (2) The reversal curse: after training on statements whose information is presented in one order (e.g., ``A is B''), arLLMs can answer questions aligned with that order but fail catastrophically when asked to invert it (e.g., ``B is A'') \citep{berglund2023reversal, zhu24b, lv-etal-2024-analysis, lin2024delving, guo-etal-2024-mitigating, golovneva2024reverse, lu2024rethinking}. Augmentation helps only when paraphrases restate the same facts in varying orders.

Recent masked diffusion language models (dLLM) \citep{ICLR2025_ce1c1ff5, llada, ye2025dream} use a bidirectional ``denoising'' style training objective, similar to their precedents such as BERT and T5/Flan-T5 \citep{devlin2019bert, raffel2020exploring}, at LLM scale. Several advantages of dLLMs have been found over arLLMs. During pretraining, dLLMs can reach lower validation loss in data-scarce regimes and generalize better to down-stream tasks than arLLMs \citep{prabhudesai2025diffusion, ni2025difflm}, suggesting better sample efficiency. dLLMs also appear less prone to the ``reversal curse''; an 8B dLLM has been reported to outperform GPT-4o on a reversal poem completion task \citep{llada}. This motivates us to study demasking as a fine-tuning objective for knowledge injection.

First, we test fine-tuning on raw text for knowledge injection and focus on evaluating QA performance in dLLMs and arLLMs. Across datasets, we find that arLLMs strongly depend on paraphrase augmentation for forward QA generalization and still fail on backward QA (an indication of the reversal curse). In contrast, dLLMs achieve high accuracy in both directions without paraphrases.

Motivated by this dLLM advantage, we ask whether the demasking objective alone is sufficient for effective knowledge injection, independently of diffusion-style decoding or bidirectional attention. If so, giving a decoder-only arLLM the demasking objective at fine-tuning time should substantially improve knowledge injection from raw text and largely close the performance gap to dLLMs. To test this hypothesis, we apply \emph{demasking fine-tuning} to arLLMs:
during fine-tuning, we create a corrupted version of the document by masking out a fraction of tokens, and prompt the model to recover the full passage. By sampling masks, a single document yields many conditioning patterns, providing bidirectional learning signals while preserving the model's decoder-only architecture and standard autoregressive training. This demasking fine-tuning objective substantially improves arLLM knowledge injection. It reduces reliance on paraphrase augmentation for forward QA and mitigates the reversal curse for backward QA, recovering most of the gap to dLLMs (Figure~\ref{fig:diagram}).

To probe whether the benefit scales beyond controlled settings\del{ and extends to procedural knowledge}, we further test the demasking objective on a large-scale knowledge-intensive dataset (Webscale-RL, 1.1M samples)\del{ and on math SFT}. On Webscale-RL, demasking-AR SFT achieves the best downstream accuracy on GPQA-diamond among all fine-tuning variants. \del{On math tasks (GSM8K and MATH), demasking SFT consistently improves over standard SFT, suggesting broad applicability of the demasking objective.}

In this paper, we make the following key contributions:
\label{sec:contributions}

\begin{enumerate}
  \item A controlled comparison of knowledge injection by fine-tuning arLLM and dLLM on three datasets shows that the dLLM learns from raw documents without paraphrases and largely escapes the reversal curse, whereas the arLLMs do not. To our knowledge, this advantage has been characterized in pre-training but not in knowledge-intensive fine-tuning.
  \item \del{We found that the demasking objective is the key ingredient rather than the diffusion architecture.} Demasking applied as an instruction-formatted fine-tuning task to arLLMs recovers most of that gap with no architectural or decoding changes.
  \item \del{We show that the demasking objective in the fine-tuning phase is a generally effective method on broad tasks such as reasoning and math.}\add{We provide preliminary evidence that the demasking objective can also help beyond controlled factual knowledge injection, on learning a large-scale knowledge-intensive corpus (Webscale-RL).}
  
\end{enumerate}

\section{Background}

\subsection{Knowledge injection by fine-tuning}

Empirically, LLMs can appear to store facts yet fail to extract them under novel question wordings unless the training samples cover a variety of ways that the same fact can be presented \citep{ovadia2023fine, mecklenburg2024injecting, gekhman2024does, soudani2024fine, zhao2025style, lampinen2025generalization}. In a synthetic setting, \citet{zhu24a} found that reliable knowledge extraction correlates strongly with paraphrase augmentation during training, and that post-hoc instruction tuning cannot fully recover extractability if the pretraining signal lacks such variation. In a more realistic setting involving Wikipedia knowledge injection into instruction-tuned LLMs, \citet{ovadia2023fine} and \citet{mecklenburg2024injecting} found that QA capability scales with the number of paraphrases used in training but saturates quickly at around 10 paraphrases per sample and still lags behind retrieval-based methods such as RAG. \citet{ovadia2025knowledge} showed that a multi-stage data augmentation pipeline---first breaking articles down into individual concepts and facts and then paraphrasing---can further improve QA accuracy, but such methods are extremely expensive and rely on a strong LLM to prepare the data.

One failure mode is the reversal curse: after learning statements of the form ``A is B,'' a model does not generalize to the inverse form ``B is A.'' It has been observed across training phases and model families \citep{berglund2023reversal, zhu24b, lv-etal-2024-analysis, lin2024delving, guo-etal-2024-mitigating, golovneva2024reverse, lu2024rethinking}. Even commercial models such as GPT-4 and GPT-4o show signs of the reversal curse \citep{berglund2023reversal, llada}. The reversal curse has been theoretically attributed to an inherent limitation of the autoregressive training objective \citep{zhu2024towards, kitouni2024factorization} (see Appendix~\ref{app:reversal} for further discussion). Common approaches to mitigating the reversal curse in autoregressive models include: (i) augmenting the training set with paraphrases \citep{lu2024rethinking}; (ii) augmenting the training set with reordered sequences \citep{guo-etal-2024-mitigating, golovneva2024reverse}, which often violates natural-language grammar and degrades overall language modeling performance; and (iii) replacing causal attention with bidirectional attention \citep{lv-etal-2024-analysis}, which struggles to retrieve information over long contexts (e.g., a person's description). In contrast, the demasking fine-tuning paradigm we study (Section~\ref{section:6}) for arLLMs addresses the reversal curse without constructing paraphrase augmentations or altering the autoregressive objective; it only requires rewriting the training sample into a demasking task.

A separate line of work injects facts by editing weights directly rather than by fine-tuning. Locate-and-edit methods such as ROME \citep{meng2022locating} and MEMIT \citep{meng2022massediting} compute a targeted update to specific MLP layers, and variants add external memories or adapters to accumulate many edits without drift \citep{hartvigsen2023aginggracelifelongmodel, yu2023meloenhancingmodelediting, wang2024wiserethinkingknowledgememory}. The two lines differ in the form the knowledge takes. Knowledge editing is formulated around atomic assertions, usually a (subject, relation, object) triple or a short statement, and the edit is specified in advance. New knowledge, however, usually arrives as unstructured documents, in which a single passage carries many interrelated facts, some of them spanning several sentences. Applying an editing method to a document therefore requires an extraction step that converts it into triples, which costs additional strong-LLM passes whose cost grows with the length of the documents \citep{wu2025unke, jiang2025anyedit}. Fine-tuning takes the document as it is, needs no extraction or target specification, and uses the same machinery as pre-training and post-training, so it composes with existing pipelines and scales to large corpora. The 1.1M-example experiment in Section~\ref{sec:scaling} would be impractical to express as individual edits.

\subsection{The demasking objective in language modeling}
\label{sec:infilling}

dLLMs have recently emerged as a strong competitor to arLLMs \citep{sahoo2024simple, llada, ye2025dream, khanna2025mercury}. Compared to autoregressive models, dLLMs use bidirectional attention to generate text by iteratively demasking tokens via a reversed discrete diffusion process. The training objective is to minimize the mask reconstruction loss \citep{llada}:

\begin{equation}
\mathcal{L}(\theta) = -\mathbb{E}_{t,\boldsymbol{x}_0,\boldsymbol{x}_t}\left[\frac{1}{t}\sum_{\ell=1}^{L}\mathbb{I}[\ell\in\mathcal{M}]\log p_\theta(\boldsymbol{x}_0^{\ell}|\boldsymbol{x}_t)\right],
\label{eq:1}
\end{equation}

$\boldsymbol{x}_0$ is an original sequence of length $L$ sampled from the training data. The masking process is governed by the mask ratio $t$, which is sampled uniformly, resulting in the corrupted sequence $\boldsymbol{x}_t$. The set $\mathcal{M}$ denotes the indices of the tokens that were masked by the forward process at ratio $t$. The $\ell$-th token is considered for the loss only if it was masked.

Recent studies report dLLMs are more sample-efficient than arLLMs. When the training data is scarce, dLLMs keep improving with repeated use of the data and surpass arLLMs on validation loss, while arLLMs saturate the validation loss or increase it due to overfitting \citep{prabhudesai2025diffusion, ni2025difflm}. \citet{prabhudesai2025diffusion} further shows that the lower validation loss in dLLMs can generalize to downstream tasks like ARC-Easy, and attributes its data efficiency to random masks as implicit data augmentation. However, whether these advantages persist in new knowledge acquisition during fine-tuning, where the model needs to learn generalizable knowledge through small fine-tuning sets, remains unclear.

Besides being used in the diffusion language models, reconstructing text from a corrupted view is a long-standing language modeling objective. BART \citep{lewis2020bart} corrupts a document with masking, deletion and span shuffling and trains a sequence-to-sequence model to restore it. ILM \citep{donahue2020ilm} casts blank-filling as a left-to-right generation task by appending the answers to the context. GLM \citep{du2022glm} performs autoregressive blank infilling over sampled spans, FIM \citep{bavarian2022efficient} moves a suffix in front of a prefix so that a causal decoder learns to fill the middle, InCoder \citep{fried2023incoder} and CM3 \citep{aghajanyan2022cm3} adopt causally-masked objectives for code and multimodal documents, and UL2 \citep{tay2023ul2} unifies these as a mixture of denoisers spanning several corruption rates. Our demasking fine-tuning for arLLMs differs from these studies in at least two axes. (1) The above infilling methods mostly remove and recover contiguous spans (e.g. FIM, GLM, ILM, and BART). We instead mask tokens independently, with each token dropped with a probability of $t$, which is the same as the dLLM forward process. (2) The works above all use the objective during pre-training, and evaluate it as a general-purpose capability. We apply it during post-training to off-the-shelf pre-trained models, on a small corpus of documents, and we do so without altering the model's loss or causal attention patterns. The corrupted document is placed in the user turn, and the original is the assistant target, so ordinary autoregressive supervised fine-tuning on a chat template implements the objective.

\section{Datasets and experimental setups}
\label{sec:setup}

We focus on assessing LLMs' ability to learn new knowledge through fine-tuning. More specifically, LLMs are fine-tuned on a set of documents that contain knowledge unknown to the base LLM, and evaluated by open-ended QA. \del{The correctness of an answer is evaluated by the ROUGE-1 recall between the generated answer and the ground truth answer, which we report as ``accuracy''.}\add{We measure the correctness of an answer in two ways: (i) \textbf{ROUGE-1 recall} \citep{lin2024delving, jiang2025anyedit} between the generated answer and the ground-truth answer; and (ii) \textbf{LLM judge} (gpt-oss-120b) that marks an answer correct if it states the reference fact or an equivalent paraphrase of it. The two measures agree on 96.6\% of 124{,}476 answer judgments (Appendix~\ref{app:judge}). Main tables report both measures; if the text quotes a single number, it is ROUGE-1 recall.} Generation examples for every condition are in Appendix~\ref{app:example}.

We use three representative datasets. Two are existing synthetic datasets from prior reversal-curse studies, each accompanied by a paraphrase set for data augmentation; the third is constructed from Wikipedia articles about recent events.
Prior work on the reversal curse has rarely used realistic datasets.
We also include a Wikipedia dataset to probe the reversal curse in a setting that better reflects real-world knowledge acquisition, where new events continually occur and models must absorb and retain them. See examples of each dataset in Appendix~\ref{app:dataset}.

The \textbf{\textit{NameDescription}} \citep{berglund2023reversal} contains 60 statements of different fictitious individuals, 30 each of the form ``[name] is [description]'' (N2D) and ``[description] is [name]'' (D2N). \cite{lin2024delving} extended the dataset with an open-ended QA testing set. For each type of statement, the QA set contains two types of questions: ``What is the name related to a given description'' and ``What is the description of a given name''. Depending on whether the question is aligned with the original statement, each question is classified as ``forward'' or ``backward'' question (e.g.\ N2D statement with ``What is the description of a given name'' type of question is a forward question). The dataset also contains a paraphrase set, in which each statement is rewritten into 30 different versions, but the order of [name] and [description] in the paraphrases is always preserved as in the original statement (either N2D or D2N).

The \textbf{\textit{Biography}} dataset is proposed in \cite{zhu24a,zhu24b}. Since the original dataset is not publicly available, we used a subset of 100 samples from a replication \citep{zheng2025spurious}. Each sample is a 6-sentence paragraph about a fictitious individual, detailing their birth city, birthday, college, and job information. Note that the name only appears in the first sentence and is replaced with a pronoun in the following sentences; thus, questions about the name are considered backward questions. Each sample also includes a paraphrase set of 5 paraphrases; the paraphrases do not change the order of the sentences but only alter the wording while preserving the information. The testing QA set has both forward (i.e., asking for an attribute given the name) and backward styles (i.e., asking for the name given 3 attributes that uniquely identify the person) questions.

The \textbf{\textit{Wiki}} dataset contains 94 Wikipedia articles, constructed according to the protocol described in \cite{pan2025memorization}. We crawl the Wikipedia pages under the category ``2025 by month'', and then further filter out the pages that were created before the year 2025. This procedure places the training data after the post-training cutoff of most of our models, and is evidenced by the models' accuracy before fine-tuning (Appendix Table~\ref{tab:3}). Because contamination is a natural concern for any Wikipedia-derived benchmark, we state the control explicitly: the pre-fine-tuning ROUGE-1 recall of every model on this dataset (\del{0.178 $\pm$ 0.049 forward, 0.190 $\pm$ 0.049 backward}\add{0.169 $\pm$ 0.062 forward, 0.199 $\pm$ 0.044 backward}, Table~\ref{tab:ar_summary}) is far below post-fine-tuning values, so the effects we report cannot be explained by prior exposure. The NameDescription and Biography datasets consist of fictitious entities and are not subject to leakage at all, and they carry the same qualitative conclusions. For each wiki article, we use o3-mini to generate QA pairs in both forward and backward styles. By prompting o3-mini, we construct two different paraphrase sets: one that retains the information in place while only changing the wording (same-order paraphrases); the other also changes the order of information in the article (permute-order paraphrases). 9 paraphrases of each type are generated for every wiki article. All LLM generated QAs or paraphrases are cross-checked by humans and invalid entries are filtered out and replaced. More details on constructing the datasets are provided in Appendix~\ref{app:dataset}.

We chose the arLLMs Llama-3.1-8B-Instruct, Llama-3.2-3B-Instruct \citep{dubey2024llama}, Qwen2.5-7B-Instruct, Qwen3-4B-Instruct-2507 \citep{yang2025qwen3} and dLLM LLaDA-8B-Instruct \citep{llada} to conduct the experiments. LLaDA-8B-Instruct is directly comparable to Llama-3.1-8B-Instruct as they have similar parameter size and comparable general capability benchmarks. \add{See Appendix~\ref{app:fine-tuning} for the choice of training hyperparameters and Appendix~\ref{app:budget} for the fixed training-compute budget at which all main results are reported.}

\section{Results}

\subsection{Failures of arLLM knowledge injection}

\begin{delblock}
\end{delblock}
\begin{addblock}
\begin{table}[h]\tablemarkup

  \caption[]{\tablemarkup arLLM fine-tuning accuracy at the fixed training-compute budget (Appendix~\ref{app:budget}), averaged across four models (Llama-3.1-8B, Llama-3.2-3B, Qwen2.5-7B, Qwen3-4B), under both accuracy measures: ROUGE-1 recall (ROUGE) and the LLM judge (Section~\ref{sec:setup}). Values are mean $\pm$ std across models; Wiki paraphrases are the same-order set. (Per-model: Appendix Tables~\ref{tab:3}, \ref{tab:judge}; peak protocol: Appendix~\ref{app:fixedstep}.)}
  \label{tab:ar_summary}
  \begin{center}
    {\fontsize{8}{9}\selectfont
        \setlength{\tabcolsep}{3pt}
        \begin{tabular}{llcccccc}
          \toprule
          & & \multicolumn{2}{c}{NameDescription} & \multicolumn{2}{c}{Biography} & \multicolumn{2}{c}{Wiki} \\
          \cmidrule(lr){3-4}\cmidrule(lr){5-6}\cmidrule(lr){7-8}
          & Metric & Fwd & Bwd & Fwd & Bwd & Fwd & Bwd \\
          \midrule
          Before fine-tuning & ROUGE & 0.027{\tiny$\pm$0.013} & 0.023{\tiny$\pm$0.009} & 0.019{\tiny$\pm$0.027} & 0.001{\tiny$\pm$0.001} & 0.169{\tiny$\pm$0.062} & 0.199{\tiny$\pm$0.044} \\
            & Judge & 0.000{\tiny$\pm$0.000} & 0.000{\tiny$\pm$0.000} & 0.000{\tiny$\pm$0.000} & 0.000{\tiny$\pm$0.000} & 0.065{\tiny$\pm$0.041} & 0.016{\tiny$\pm$0.013} \\
          \addlinespace
          w/o paraphrases & ROUGE & 0.257{\tiny$\pm$0.169} & 0.025{\tiny$\pm$0.008} & 0.106{\tiny$\pm$0.125} & 0.001{\tiny$\pm$0.001} & 0.478{\tiny$\pm$0.154} & 0.321{\tiny$\pm$0.095} \\
            & Judge & 0.231{\tiny$\pm$0.173} & 0.009{\tiny$\pm$0.009} & 0.067{\tiny$\pm$0.090} & 0.000{\tiny$\pm$0.000} & 0.401{\tiny$\pm$0.159} & 0.165{\tiny$\pm$0.071} \\
          \addlinespace
          w/ paraphrases & ROUGE & 0.925{\tiny$\pm$0.047} & 0.042{\tiny$\pm$0.013} & 0.881{\tiny$\pm$0.070} & 0.002{\tiny$\pm$0.002} & 0.630{\tiny$\pm$0.063} & 0.379{\tiny$\pm$0.022} \\
            & Judge & 0.926{\tiny$\pm$0.055} & 0.017{\tiny$\pm$0.013} & 0.859{\tiny$\pm$0.081} & 0.000{\tiny$\pm$0.000} & 0.588{\tiny$\pm$0.045} & 0.225{\tiny$\pm$0.034} \\
          \bottomrule
        \end{tabular}
    }
  \end{center}
\end{table}
\end{addblock}

We first show that knowledge injection by fine-tuning in arLLMs heavily relies on paraphrases. This is known in previous studies \citep{berglund2023reversal, zhu24b, lin2024delving, guo-etal-2024-mitigating, golovneva2024reverse}. We demonstrate this observation on three datasets to set baselines for comparison with dLLM and demasking fine-tuning in the following sections.

We fine-tune arLLMs on samples from each dataset using a chat template format, in which a generic instruction (i.e., ``Tell me a fact.'') is followed by the knowledge text as the target completion. Training without the chat template (i.e., continued pre-training format) yields qualitatively similar conclusions (Appendix~\ref{app:chat_template}). Without paraphrases, backward accuracy on the NameDescription and Biography datasets is close to 0, while the forward accuracy of NameDescription N2D and Biography does not completely fail but is still poor (Table~\ref{tab:ar_summary}). Adding same-order paraphrases drastically raises forward accuracy close to 1, while backward accuracy remains close to 0. Paraphrases do not help backward accuracy in NameDescription and Biography datasets because the construction of these datasets does not change the semantic order of the sentences.
The trend is similar in the Wiki dataset (Table~\ref{tab:ar_summary}). While the same-order paraphrases significantly increase forward accuracy, they only mildly increase backward accuracy. Using permute-order paraphrases increases both forward and backward accuracy, and the gap between them is smaller (Appendix Figure~\ref{fig:wiki_ar}). Note that, due to the naturalness of this dataset, pre-fine-tuning accuracies are not as close to zero as in the other datasets; nonetheless, they are sufficiently low to demonstrate the effectiveness of fine-tuning.

These results suggest that, in arLLM fine-tuning, paraphrases significantly improve QA performance, but help backward questions only when the paraphrases change the information order in the original text to be more aligned with the backward style. \del{The difference between fine-tuning with and without paraphrases is not due to different training steps: both are trained for a sufficiently large number of epochs and each is reported at its own best checkpoint, and the same conclusion holds at a shared fixed budget (Appendix~\ref{app:fixedstep})}\add{By construction, fine-tuning with and without paraphrases is compared at equal training compute, so the difference is not due to different training amounts; the same conclusion holds at every other budget on the grid and at the peak protocol (Appendix~\ref{app:fixedstep})} (training curves in Appendix Figure~\ref{fig:main} and Appendix Figure~\ref{fig:total_acc}).

\subsection{Effectiveness of dLLM knowledge injection}

\begin{addblock}
\begin{table}[h]\tablemarkup

  \caption[]{\tablemarkup dLLM (LLaDA-8B-Instruct) fine-tuning accuracy at the same compute budgets as Table~\ref{tab:ar_summary}, under both accuracy measures. (Per-model: Appendix Table~\ref{tab:3}.)}
  \label{tab:dllm_summary}
  \begin{center}
    {\fontsize{8}{9}\selectfont
        \setlength{\tabcolsep}{3pt}
        \begin{tabular}{llcccccc}
          \toprule
          & & \multicolumn{2}{c}{NameDescription} & \multicolumn{2}{c}{Biography} & \multicolumn{2}{c}{Wiki} \\
          \cmidrule(lr){3-4}\cmidrule(lr){5-6}\cmidrule(lr){7-8}
          & Metric & Fwd & Bwd & Fwd & Bwd & Fwd & Bwd \\
          \midrule
          Before fine-tuning & ROUGE & 0.015 & 0.014 & 0.030 & 0.000 & 0.156 & 0.210 \\
            & Judge & 0.000 & 0.000 & 0.001 & 0.000 & 0.078 & 0.050 \\
          \addlinespace
          w/o paraphrases & ROUGE & 0.888 & 0.769 & 0.847 & 0.345 & 0.922 & 0.694 \\
            & Judge & 0.868 & 0.744 & 0.605 & 0.214 & 0.862 & 0.612 \\
          \addlinespace
          w/ paraphrases & ROUGE & 0.948 & 0.927 & 0.979 & 0.683 & 0.899 & 0.755 \\
            & Judge & 0.930 & 0.908 & 0.960 & 0.720 & 0.853 & 0.663 \\
          \bottomrule
        \end{tabular}
    }
  \end{center}
\end{table}
\end{addblock}

Inspired by the known advantages of dLLMs \citep{prabhudesai2025diffusion, ni2025difflm, llada}, we investigate whether they require paraphrases during fine-tuning to successfully handle both forward and backward QAs. We follow the original pretraining protocol \citep{llada} to fine-tune LLaDA-8B-Instruct on the dataset samples using the loss defined in Eq.~\ref{eq:1}. On NameDescription and Wiki, the accuracy difference between fine-tuning with and without paraphrases is much smaller in the dLLM than in the arLLM (Table~\ref{tab:dllm_summary}): dLLM without paraphrases can already achieve decent and comparable accuracies on both forward and backward questions\add{; on Biography backward QA the dLLM converges slowly (Appendix Table~\ref{tab:convergence}), reaching 0.35 at the budget but 0.70 at its peak (Appendix~\ref{app:fixedstep})}; fine-tuning with paraphrases can further increase the accuracy by a small amount. Taken together, these results suggest that the dLLM is substantially less reliant on paraphrase augmentation during post-training and exhibits markedly reduced reversal-curse behavior in this setting. By plotting the test accuracy across the training steps (Appendix Figure~\ref{fig:main}), we observe that arLLM fine-tuned without paraphrases improves QA accuracy only in the beginning of training, then quickly decreases, indicating overfitting. The dLLM without paraphrases, on the other hand, does not show signs of overfitting. This finding echoes what has been found in comparing arLLMs and dLLMs in the pre-training phase \citep{prabhudesai2025diffusion, ni2025difflm}.

One may expect that fine-tuning dLLM converges slower than arLLM, because learning any-order factorization requires seeing more than one way of the factorizations (i.e., samples masked in different ways) \citep{xue2025any, kim2025train}. However, apart from Biography backward QA (see above), we do not observe the dLLM needing substantially more optimization steps to reach its plateau (Appendix Figure~\ref{fig:main}). To control for differences in instruction-following capability across models, we additionally evaluated all fine-tuned models using few-shot (2-shot) prompting (Appendix~\ref{app:fewshot}). The  conclusion is qualitatively unchanged.

\subsection{Demasking fine-tuning of arLLM}
\label{section:6}

\begin{figure}[h]
\centering
\fbox{\parbox{0.75\linewidth}{\small
\textbf{User:}\ \ {[MASK]} Barrington, known {[MASK]} and {[MASK]} for being {[MASK]} acclaimed director of the {[MASK]} reality masterpiece, ``A {[MASK]} Through {[MASK]}.'' \textbackslash n Return the recovered masked passage.

\medskip
\textbf{Assistant:}\ \ Here is the recovered text:\textbackslash n \textcolor{BrickRed}{Daphne Barrington, known far and wide for being the acclaimed director of the virtual reality masterpiece, ``A Journey Through Time.''}
}}
\caption{An example of demasking fine-tuning prompt. A random selection of tokens is replaced by [MASK] token. Red tokens are used to compute the autoregressive loss.}
\label{box}
\end{figure}

Motivated by the dLLM results above, we ask whether the same advantage can be obtained by an arLLM. If an instruct arLLM is capable enough, one may prompt an arLLM to act like a dLLM (Figure~\ref{box}). Specifically, given a masked document and an instruction to recover the original document, if the model has knowledge of the original document and the masked document contains sufficient cues to retrieve this knowledge, an instruct arLLM should respond with the correct original document. If the arLLM does not already have the knowledge of the original document, setting the ground truth document as the supervised fine-tuning target may teach the model that knowledge. We refer to this fine-tuning paradigm as ``\textit{demasking fine-tuning}'' of arLLM, and the resulting model as ``\textit{demasking arLLM}.'' Demasking fine-tuning of arLLM, from a broader perspective, establishes a training objective similar to that of dLLMs, wherein the model learns to reconstruct the unmasked sequence from a masked input. Following the dLLM noise sampling strategy, we randomly replace sample tokens with a reserved special token during training, where the mask ratio \(t \) is sampled from a uniform distribution $U(0.05,0.95)$. Each sample can be masked with a different mask ratio in different epochs. We evaluate the demasking fine-tuned arLLM in the regular autoregressive way using the default chat template. The exact prompt used in the fine-tuning is provided in Figure~\ref{box} (more details in Appendix~\ref{app:fine-tuning}). The training objective is:

\begin{equation}
\mathcal{L}(\theta) = - \frac{1}{\sum m_{i}}
\sum_{i=1}^{T} m_{i}\, \log p_{\theta}\!\big(s_{i}\,\big|\,s_{< i}\big)
\label{eq:2}
\end{equation}

where $s$ is the constructed sequence which has the form of Figure~\ref{box}; $m_i\in\{0,1\}$ selects which tokens are included in the loss which are 1 if the token is part of the original sample in the ``assistant'' window or it is an ``end of sequence'' token.

\begin{figure}[h]
  \centering
  \includegraphics[width=\textwidth]{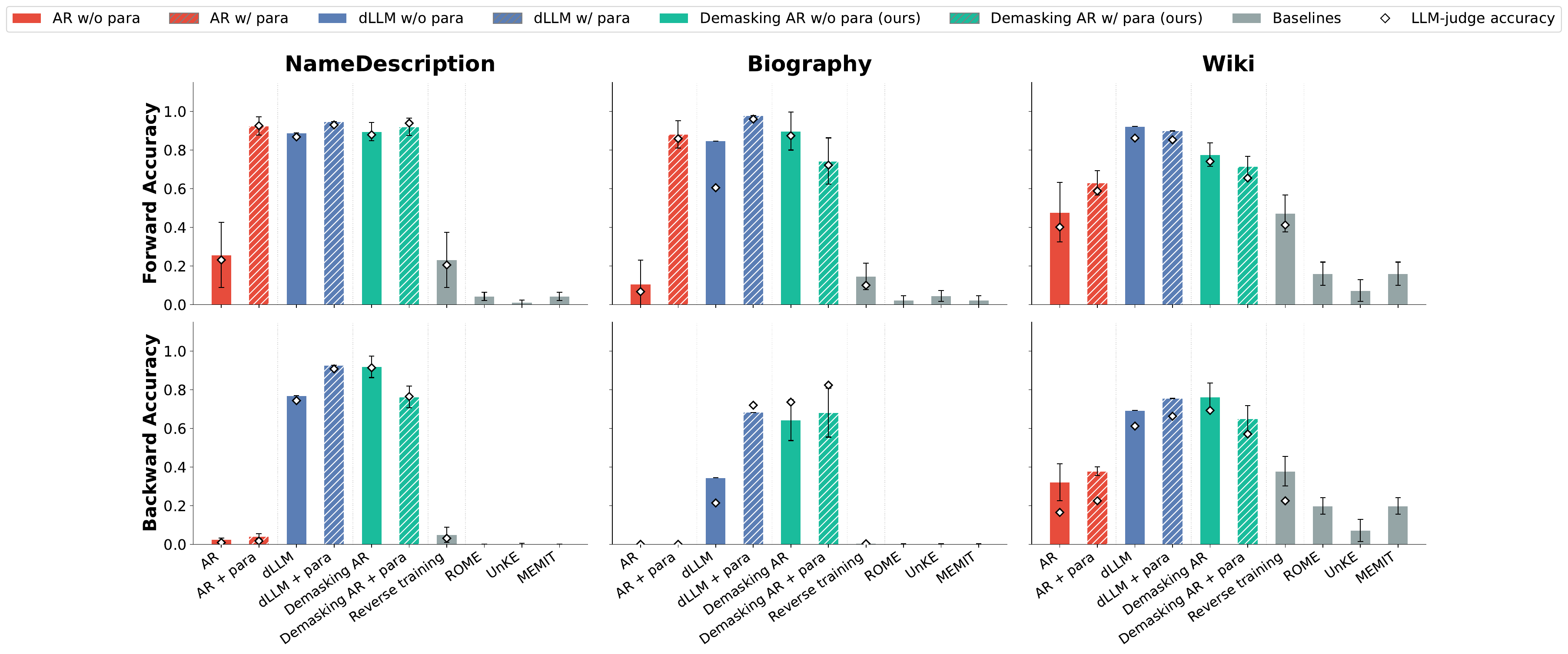}
  \caption{Summary of QA accuracy \add{at the fixed training-compute budget (Appendix~\ref{app:budget}; equal total FLOPs across methods)} across all models (Llama-3.1-8B, Llama-3.2-3B, Qwen2.5-7B, Qwen3-4B), averaged per method. Error bars indicate standard deviation across models \add{(dLLM: single model)}. Hatched bars denote training with paraphrase augmentation\add{; bars show ROUGE-1 recall and white diamonds the LLM-judge accuracy at the same checkpoints (knowledge-editing baselines: ROUGE-1 only; their judge values are likewise ${\approx}0$)}. Colors denote the training method: arLLM (red), dLLM (blue), demasking arLLM (teal) and knowledge-editing baselines (gray). \add{(Per-model: Appendix Table~\ref{tab:3}; peak protocol: Appendix~\ref{app:fixedstep}.)}}
  \label{fig:summary}
\end{figure}

\begin{addblock}
\begin{table}[h]\tablemarkup
  \caption[]{\tablemarkup Demasking fine-tuning accuracy (without paraphrases) with baselines on the three datasets, at the fixed training-compute budget (Appendix~\ref{app:budget}). Values are mean $\pm$ std across the four arLLMs; the dLLM is a single model (LLaDA-8B-Instruct) and therefore has no across-model spread. (Knowledge-editing baselines: Appendix~\ref{app:knowledge_editing}.)}
  \label{tab:baselines}
  \begin{center}
    {\fontsize{8}{9}\selectfont
    \setlength{\tabcolsep}{4pt}
    \begin{tabular}{llcccccc}
      \toprule
      & & \multicolumn{2}{c}{NameDescription} & \multicolumn{2}{c}{Biography} & \multicolumn{2}{c}{Wiki} \\
      \cmidrule(lr){3-4}\cmidrule(lr){5-6}\cmidrule(lr){7-8}
      Method & Metric & Fwd & Bwd & Fwd & Bwd & Fwd & Bwd \\
      \midrule
      Demasking (ours) & ROUGE & 0.895{\tiny$\pm$0.047} & 0.919{\tiny$\pm$0.056} & 0.898{\tiny$\pm$0.098} & 0.643{\tiny$\pm$0.106} & 0.778{\tiny$\pm$0.061} & 0.762{\tiny$\pm$0.072} \\
       & Judge & 0.879{\tiny$\pm$0.057} & 0.914{\tiny$\pm$0.061} & 0.874{\tiny$\pm$0.108} & 0.736{\tiny$\pm$0.127} & 0.741{\tiny$\pm$0.082} & 0.693{\tiny$\pm$0.058} \\
      \midrule
      AR fine-tuning & ROUGE & 0.257{\tiny$\pm$0.169} & 0.025{\tiny$\pm$0.008} & 0.106{\tiny$\pm$0.125} & 0.001{\tiny$\pm$0.001} & 0.478{\tiny$\pm$0.154} & 0.321{\tiny$\pm$0.095} \\
       & Judge & 0.231{\tiny$\pm$0.173} & 0.009{\tiny$\pm$0.009} & 0.067{\tiny$\pm$0.090} & 0.000{\tiny$\pm$0.000} & 0.401{\tiny$\pm$0.159} & 0.165{\tiny$\pm$0.071} \\
      \addlinespace
      Reverse training & ROUGE & 0.231{\tiny$\pm$0.143} & 0.049{\tiny$\pm$0.039} & 0.146{\tiny$\pm$0.069} & 0.006{\tiny$\pm$0.005} & 0.472{\tiny$\pm$0.096} & 0.379{\tiny$\pm$0.076} \\
       & Judge & 0.205{\tiny$\pm$0.154} & 0.031{\tiny$\pm$0.042} & 0.100{\tiny$\pm$0.063} & 0.003{\tiny$\pm$0.003} & 0.412{\tiny$\pm$0.120} & 0.225{\tiny$\pm$0.058} \\
      \addlinespace
      Prompt distillation & ROUGE & 0.607{\tiny$\pm$0.105} & 0.643{\tiny$\pm$0.084} & 0.497{\tiny$\pm$0.107} & 0.002{\tiny$\pm$0.003} & 0.613{\tiny$\pm$0.097} & 0.455{\tiny$\pm$0.100} \\
       & Judge & 0.491{\tiny$\pm$0.115} & 0.511{\tiny$\pm$0.188} & 0.451{\tiny$\pm$0.113} & 0.001{\tiny$\pm$0.002} & 0.552{\tiny$\pm$0.097} & 0.329{\tiny$\pm$0.116} \\
      \addlinespace
      dLLM & ROUGE & 0.888 & 0.769 & 0.847 & 0.345 & 0.922 & 0.694 \\
       & Judge & 0.868 & 0.744 & 0.605 & 0.214 & 0.862 & 0.612 \\
      \addlinespace
      RAG & ROUGE & 0.943{\tiny$\pm$0.038} & 0.929{\tiny$\pm$0.023} & 0.943{\tiny$\pm$0.040} & 0.934{\tiny$\pm$0.003} & 0.978{\tiny$\pm$0.005} & 0.940{\tiny$\pm$0.025} \\
       & Judge & 0.933{\tiny$\pm$0.024} & 0.944{\tiny$\pm$0.016} & 0.919{\tiny$\pm$0.049} & 0.929{\tiny$\pm$0.003} & 0.985{\tiny$\pm$0.008} & 0.947{\tiny$\pm$0.022} \\
      \bottomrule
    \end{tabular}}
  \end{center}
\end{table}
\end{addblock}

Overall, demasking fine-tuning of arLLM successfully reproduces the benefit observed in the dLLM fine-tuning (Figure~\ref{fig:summary}; Appendix Table~\ref{tab:3}, Appendix Figure~\ref{fig:main}). Demasking arLLM surpasses arLLM fine-tuning in the pre-training style by a large margin and achieves comparably high accuracy in both forward and backward question categories. Moreover, like dLLM, demasking arLLM relies much less on paraphrases in the fine-tuning dataset to saturate the accuracy in most cases. \add{On the Wiki dataset, adding paraphrases in fact \emph{lowers} the measured demasking accuracy. Appendix~\ref{app:wiki_para} analyzes this reversal and shows that it reflects a difference in response style, not in injected knowledge; few-shot prompting that fixes the style closes the gap.}

Demasking fine-tuning roughly doubles per-step FLOPs, because both the masked and the original sequence are present in the training example, and it raises peak memory correspondingly. Against this it requires no paraphrase generation, which for the Wiki dataset would otherwise cost roughly 0.1M tokens from a strong LLM, and the cost grows with the document length since conquering the reversal curse requires exploiting many permutations of the concepts in the document. We report per-step FLOPs, wall time and memory in Appendix~\ref{app:compute}. Training standard fine-tuning without paraphrases longer does not help: its accuracy decays toward zero rather than improving (Appendix~\ref{app:fixedstep}, Appendix Figure~\ref{fig:main}). \add{Note that the main-table comparison holds demasking to equal total training FLOPs by halving its passes; giving demasking the same number of passes as standard fine-tuning (${\approx}2\times$ the FLOPs) yields similar or slightly higher accuracy (Appendix~\ref{app:budget}). Within the same budget, demasking reaches 0.90/0.92 forward/backward on NameDescription while standard fine-tuning reaches 0.26/0.03. It also needs no paraphrase generation, so it adds no data-preparation compute. Most importantly, backward questions are solved by no cheaper alternative: standard fine-tuning fails them at every budget we tested, with or without paraphrases, and reverse training and knowledge editing do not close the gap either.}

To confirm that the effectiveness of demasking fine-tuning stems from the objective itself and not merely from a simple data augmentation or templating effect (i.e., introducing varied input text prepended to the training sample target), we conducted a control experiment in which the masked document in the prompt was replaced entirely with random tokens (Appendix Figure~\ref{fig:random_augmentation}). This substitution caused the accuracy of the demasking fine-tuning to drop to the level of naive arLLM fine-tuning. \add{We also compared auto-regressive fine-tuning in a chat template to the raw continued-pre-training format, where the performance difference is tiny (a mean difference of $+0.04$) (Appendix~\ref{app:chat_template}), so the instruction format alone cannot explain the demasking advantage either. Other components in our demasking training are hard to control separately and to credit for the performance gain, e.g. the duplication of each sample into a corrupted and a clean view is unavoidable in any autoregressive implementation of the demasking objective. We claim that the demasking-style training recipe is \emph{sufficient} to recover most of the dLLM advantage, but we do not claim that all components of this particular recipe are necessary for successful knowledge injection: a training recipe that allows information to flow from later tokens to earlier ones and presents naturally varying views of a single training sample (e.g., different random masks or different infilled spans) may work equally well.} 

\begin{addblock}
To position demasking fine-tuning among knowledge-injection approaches beyond standard fine-tuning, we contrast it with three types of additional methods (Table~\ref{tab:baselines}; averages over the four arLLMs, all without paraphrase augmentation; implementation and compute-matching details in Appendix~\ref{app:baselines}). (i) \textbf{Knowledge editing}: three representative baselines (ROME \citep{meng2022locating}, MEMIT \citep{meng2022massediting}, and UnKE \citep{wu2025unke}) all leave accuracy at or near the pre-fine-tuning level on every model and dataset (Appendix~\ref{app:knowledge_editing}), implying such methods are not suitable for document-level knowledge injection. (ii) \textbf{Prompt distillation} \citep{kujanpaa2024efficient}: a self-distillation method that generates QA pairs about each document with the model itself and distills the document-conditioned teacher's token distributions into a LoRA student without the document in context (rank 1024 for the 3B/4B models and 512 for the 7B/8B models; the original paper used rank 1024 for 3B and 8B and 512 for 14B). It lands between standard fine-tuning and demasking: it clearly outperforms naive fine-tuning (e.g., NameDescription backward 0.64 vs.\ 0.03) because the teacher's document-conditioned distributions carry knowledge into the student, but it inherits the forward bias of its self-generated questions and fails completely on Biography backward QA (0.002), where answering requires aggregating three attributes that its generated questions never combine. (iii) \textbf{RAG}: retrieving the top-4 documents by embedding similarity (bge-large-en-v1.5) and answering in context with the base instruct models achieves near-perfect accuracy in both directions (retrieval hit-rate 0.97--1.00).

RAG-based systems are known to be excellent at supplying new knowledge in context and to outperform fine-tuning at this task \citep{ovadia2023fine, soudani2024fine}, whereas in-weights learning is a much harder problem. This work aims to narrow the gap between in-weights learning and the retrieval ceiling, not to replace retrieval in today's use cases.
\end{addblock}

\subsection{Effects of fine-tuning mask ratio}
\label{sec:maskratio}

Previous studies \citep{zhu24a, zhu24b} claim that bidirectional BERT-like models struggle with even forward style knowledge extraction due to the mask loss, which causes the model to learn incorrect associations between tokens. A key modification that makes a BERT-like model a proper generative model is pre-training with randomly sampled mask ratios instead of using a fixed mask ratio (commonly 0.15 in BERT) \citep{llada, devlin2019bert}. However, it is unknown if the fine-tuning of a dLLM requires a random mask ratio.

\begin{figure}[h]
  \centering
  \includegraphics[width=0.8\linewidth]{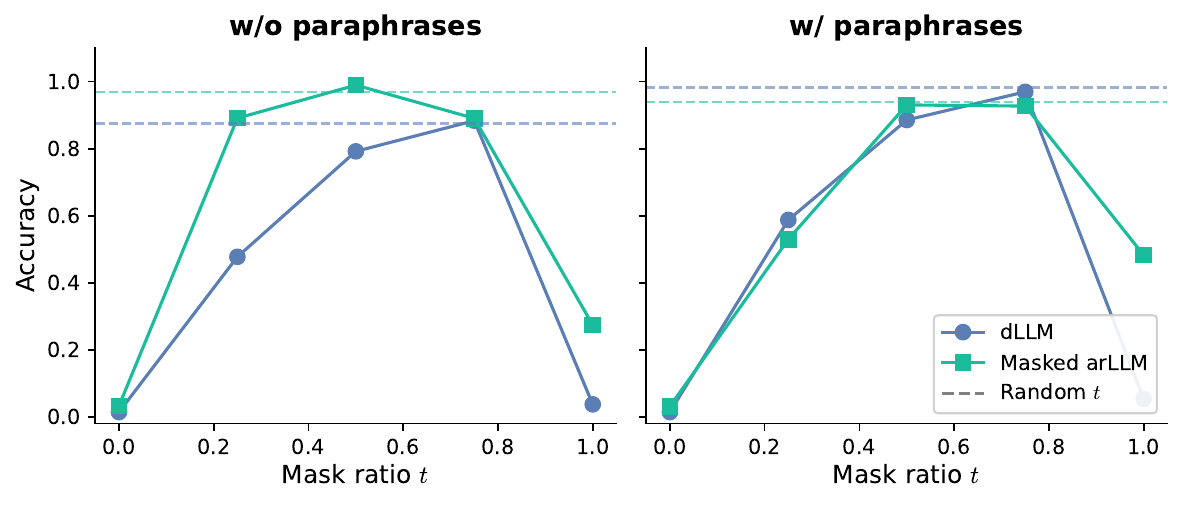}
  \caption{Peak QA accuracy on NameDescription for different fixed mask ratios ($t$). Accuracy is the average of forward and backward accuracy, taken at the best training epoch. Dashed lines show the random-$t$ baseline $U(0.05,0.95)$. The dLLM has no trainable loss at $t{=}0$; its value there shows the base model accuracy before fine-tuning.}
  \label{fig:fix_t_line}
\end{figure}

To investigate whether this necessity persists during post-training, we change the fine-tuning process of the dLLMs and demasking arLLMs to use fixed mask ratios ($t$) instead of randomly sampling them during the training (Figure~\ref{fig:fix_t_line}). Fine-tuning with some fixed mask ratios (0.75 and 0.5) can be as effective as the random mask ratio in knowledge injection. However, there is considerable performance variation across the choices of $t$.
Conceptually, varying $t$ modulates task difficulty; a moderate $t$ places training in a ``hard but not impossible'' regime that maximizes the per-example gradient signal. The effectiveness of some fixed mask ratios indicates that a dLLM only needs to vary $t$ to learn the demasking process during pre-training. Once this ability is acquired and not forgotten during fine-tuning, a fixed $t$ is sufficient for learning new knowledge.

Using a mask ratio of 0 in the demasking fine-tuning of arLLM completely fails (Figure~\ref{fig:fix_t_line}): the sample is completely exposed in the prompt with no masks, so the reconstruction task becomes trivial and yields no learning signal. Full training dynamics are shown in Appendix Figure~\ref{fig:fix_t}.

\subsection{\texorpdfstring{\add{Catastrophic forgetting}}{Catastrophic forgetting}}
\label{sec:forgetting}

Knowledge injection is only useful if the model retains what it already knew. We evaluate every fine-tuning condition on MMLU, GSM8K and IFEval (Table~\ref{tab:budget_forget}; evaluation details in Appendix~\ref{app:budget}). \add{The capability cost is modest (MMLU within $-0.03$ for nearly all conditions; Table~\ref{tab:budget_forget}). Moreover, demasking forgets less than standard fine-tuning: at the FLOP-matched checkpoints its drop is smaller in 19 of the 21 setting--benchmark comparisons (one tie; the one exception is IFEval on Biography without paraphrases). For example, on NameDescription with paraphrases standard fine-tuning loses $0.09/0.36/0.49$ of MMLU/GSM8K/IFEval versus $0.02/0.12/0.17$ for demasking---and demasking does not need paraphrases in the first place.} There are known methods to mitigate such catastrophic forgetting further, such as using LoRA \citep{hu2022lora}, rehearsal on pretraining data \citep{huang2024mitigating}, and on-policy distillation \citep{agarwal2024policy}. We expect these methods to help retain the model general capability, leaving it for future work to explore combinations of the training objectives and forgetting-mitigating methods.

\begin{table}[h]\tablemarkup

  \caption[]{\tablemarkup General-capability change relative to the base model ($\Delta$ = fine-tuned $-$ base, mean$\pm$std over the four arLLMs), measured at the same checkpoints that produce the main results. (Protocol: Appendix~\ref{app:budget}.)}
  \label{tab:budget_forget}
  \begin{center}{\fontsize{7}{8.5}\selectfont\setlength{\tabcolsep}{4pt}
  \begin{tabular}{llccc}
  \toprule
  Setting & Condition & $\Delta$MMLU & $\Delta$GSM8K & $\Delta$IFEval \\
  \midrule
  NameDescription w/o para & AR & -0.03{\tiny$\pm$0.04} & -0.20{\tiny$\pm$0.29} & -0.26{\tiny$\pm$0.24} \\
  NameDescription w/o para & Demask & -0.03{\tiny$\pm$0.03} & -0.11{\tiny$\pm$0.07} & -0.25{\tiny$\pm$0.05} \\
  NameDescription w/ para & AR & -0.09{\tiny$\pm$0.06} & -0.36{\tiny$\pm$0.33} & -0.49{\tiny$\pm$0.07} \\
  NameDescription w/ para & Demask & -0.02{\tiny$\pm$0.01} & -0.12{\tiny$\pm$0.07} & -0.17{\tiny$\pm$0.02} \\
  \midrule
  Biography w/o para & AR & -0.06{\tiny$\pm$0.07} & -0.24{\tiny$\pm$0.35} & -0.10{\tiny$\pm$0.14} \\
  Biography w/o para & Demask & -0.02{\tiny$\pm$0.01} & -0.19{\tiny$\pm$0.04} & -0.20{\tiny$\pm$0.05} \\
  Biography w/ para & AR & -0.02{\tiny$\pm$0.03} & -0.27{\tiny$\pm$0.13} & -0.26{\tiny$\pm$0.17} \\
  Biography w/ para & Demask & -0.01{\tiny$\pm$0.01} & -0.21{\tiny$\pm$0.05} & -0.14{\tiny$\pm$0.02} \\
  \midrule
  Wiki w/o para & AR & -0.03{\tiny$\pm$0.03} & -0.22{\tiny$\pm$0.10} & -0.13{\tiny$\pm$0.13} \\
  Wiki w/o para & Demask & -0.01{\tiny$\pm$0.01} & -0.08{\tiny$\pm$0.02} & -0.09{\tiny$\pm$0.03} \\
  Wiki w/ para & AR & -0.02{\tiny$\pm$0.02} & -0.18{\tiny$\pm$0.10} & -0.15{\tiny$\pm$0.13} \\
  Wiki w/ para & Demask & -0.01{\tiny$\pm$0.00} & -0.08{\tiny$\pm$0.01} & -0.06{\tiny$\pm$0.02} \\
  Wiki w/ permute para & AR & -0.02{\tiny$\pm$0.01} & -0.20{\tiny$\pm$0.12} & -0.17{\tiny$\pm$0.12} \\
  Wiki w/ permute para & Demask & -0.01{\tiny$\pm$0.00} & -0.09{\tiny$\pm$0.03} & -0.05{\tiny$\pm$0.04} \\
  \bottomrule
  \end{tabular}}\end{center}
\end{table}

\subsection{Large-scale knowledge injection with Webscale-RL}
\label{sec:scaling}

To study whether the advantages of demasking fine-tuning over regular fine-tuning of arLLM extend beyond controlled factual knowledge injection, we test this paradigm at scale on a large, realistic knowledge-intensive dataset. We fine-tune Qwen3-4B on the Webscale-RL dataset \citep{webscalerl} (details in Appendix~\ref{app:sft}), which contains \del{1.2}\add{1.1} million high-quality knowledge-intensive examples across more than 9 domains. Each training example comes in two forms: unstructured raw text (continued pre-training, CPT) and constructed QA pairs (SFT); CPT subsamples every third example after length filtering (Appendix~\ref{app:sft}), so the 1.1M figure applies to SFT, not to the examples actually used for CPT. We train four models: CPT AR, CPT demasking-AR, SFT AR, and SFT demasking-AR, and evaluate on GPQA-diamond \citep{rein2024gpqa}, a challenging graduate-level science QA benchmark. \del{Demasking-AR SFT achieves the highest GPQA-diamond accuracy (0.389), outperforming standard SFT AR (0.359) and the base model (0.242). This demonstrates that the demasking objective scales to realistic, large-scale knowledge-intensive training.}\add{We split GPQA-diamond into a 99-question validation set, used only to pick the training checkpoint per variant, and a 99-question test set on which we report accuracy with bootstrap confidence intervals (Table~\ref{tab:webscale}). Under this protocol, demasking-AR SFT attains the best test accuracy (0.394, vs.\ 0.323 for standard SFT AR and 0.273 for the base model). The same ordering holds for CPT (0.333 vs.\ 0.293). In a paired bootstrap analysis, demasking exceeds standard fine-tuning in 88\% of resamples for SFT and 87\% for CPT, and exceeds the base model in 97\%, though the two objective comparisons are not individually significant at this benchmark size (Appendix~\ref{app:sft}). We read this experiment as promising evidence that the demasking objective helps at realistic corpus scale on one base model and one benchmark, not as a conclusive superiority claim. Beyond the target benchmark, Table~\ref{tab:webscale_bench} shows that at its selected checkpoint demasking-AR SFT fully preserves general capability (MMLU 0.710 vs.\ base 0.706; IFEval 0.691 = base), whereas standard SFT AR loses 9.6 points of IFEval; all fine-tuned variants improve BBH over the base model, though part of that gain reflects improved compliance with the benchmark's answer-extraction format rather than capability.}

\begin{addblock}
\begin{table}[h]\tablemarkup

  \caption[]{\tablemarkup Large-scale fine-tuning on Webscale-RL (1.1M samples), Qwen3-4B on GPQA-diamond with a 99/99 validation/test split. Checkpoints are selected on the validation half only; we report test accuracy with 95\% bootstrap confidence intervals, and the peak accuracy on the full benchmark for comparability with the peak-reporting protocol of Appendix~\ref{app:fixedstep}.}
  \label{tab:webscale}
  \vspace{2pt}
  \centering
  {\fontsize{8}{9}\selectfont
    \setlength{\tabcolsep}{4pt}
    \begin{tabular}{lccc}
      \toprule
      & Val (selection) & Test [95\% CI] & Full-set peak \\
      \midrule
      Qwen3-4B (base)   & 0.212 & 0.273 [0.192, 0.364] & 0.242 \\
      CPT AR            & 0.263 & 0.293 [0.212, 0.384] & 0.278 \\
      CPT demasking-AR  & 0.293 & 0.333 [0.242, 0.424] & 0.313 \\
      SFT AR            & 0.394 & 0.323 [0.232, 0.424] & 0.359 \\
      SFT demasking-AR  & \textbf{0.424} & \textbf{0.394} [0.293, 0.485] & \textbf{0.409} \\
      \bottomrule
    \end{tabular}
  }
\end{table}

\begin{table}[h]\tablemarkup

  \caption[]{\tablemarkup General-capability benchmarks of the Webscale-RL fine-tuned models at their validation-selected checkpoints (MMLU 0-shot acc; IFEval instruction-level strict; BBH 3-shot exact match).}
  \label{tab:webscale_bench}
  \vspace{2pt}
  \centering
  {\fontsize{8}{9}\selectfont
    \setlength{\tabcolsep}{4pt}
    \begin{tabular}{lccc}
      \toprule
      & MMLU & IFEval & BBH \\
      \midrule
      Qwen3-4B (base)   & 0.706 & \textbf{0.691} & 0.307 \\
      CPT AR            & 0.710 & 0.483 & \textbf{0.698} \\
      CPT demasking-AR  & 0.711 & 0.655 & 0.617 \\
      SFT AR            & \textbf{0.716} & 0.595 & 0.455 \\
      SFT demasking-AR  & 0.710 & \textbf{0.691} & 0.488 \\
      \bottomrule
    \end{tabular}
  }
\end{table}
\end{addblock}

An intuition for why demasking SFT \del{helps in problem-solving settings}\add{can help in knowledge-intensive SFT settings such as Webscale-RL} is that the objective acts as a form of implicit curriculum: the model must complete the response from varying starting points, filling in the ``missing links'' of a partial answer rather than only predicting the next token of a fixed linear sequence. Reconstructing omitted spans from the surrounding context trains the model on the dependencies within a full response, making it less dependent on any specific prefix to reach the correct answer.

\section{Discussion}


We believe knowledge injection via fine-tuning will be central to self-evolving AI in the era of experience \citep{silver2025welcome}. Most current memory systems rely on external databases storing knowledge as text. Such explicit memory leverages in-context learning but suffers from context window limitations \citep{liu2023lost}, computational cost of long contexts, inability to express implicit non-semantic knowledge, and retrieval bottlenecks in vector embeddings \citep{weller2025theoretical}\add{, as well as the need to already know what to search for before searching \citep{xiao2026memory}}. Parametric memory avoids these issues but remains underused in practice due to the complications of fine-tuning \citep{zhang2025survey}.
Our results suggest that the demasking objective provides a potential route to reliable parametric knowledge updates in autoregressive LLMs. We hope future work scales this paradigm to continual, real-world memory and agent learning settings.

\begin{addblock}
\section*{Broader Impact Statement}
This work studies techniques for injecting factual knowledge into model parameters by fine-tuning on raw documents. Beneficial uses include keeping deployed models factually current without expensive paraphrase-augmentation pipelines. The same techniques, however, could make it easier to insert incorrect, biased, private, or deliberately misleading content into a model's weights, where such edits are less transparent and harder to audit or roll back than retrieval-based memory. Deployments that adopt parametric knowledge updates should therefore pair them with provenance tracking of training documents, evaluation against misinformation and privacy leakage, monitoring of general-capability regressions, and rollback mechanisms. Our experiments also show that aggressive fine-tuning degrades general capabilities; practitioners should monitor this trade-off explicitly.
\end{addblock}

\subsubsection*{Acknowledgments}
This work was supported by the Kempner Institute for the Study of Natural and Artificial Intelligence at Harvard University, the Office of Naval Research (ONR) grant No.\ N00014-23-1-2051, the Gatsby Charitable Foundation, and an Amazon Research Award Spring 2025. Any opinions, findings, and conclusions or recommendations expressed in this material are those of the authors and do not reflect the views of Amazon. We have benefited from helpful discussions with Jorin Overwiening and Binxu Wang.

\bibliographystyle{tmlr}
\bibliography{references}

\newpage
\appendix

\section{Appendix}


\subsection{Dataset and code availability}
\label{sec:code}

The dataset and codebase are available at \url{https://github.com/xup5/masked_arLLM}.


\subsection{LLM usage}

LLM is used for editing (e.g., grammar, spelling, word choice), drafting sections of the paper, Data processing/filtering, Visualizing results for submission, Facilitating or running experiments, and implementing standard methods. The authors checked the correctness of LLM generated content.


\subsection{Dataset details and examples}
\label{app:dataset}

All the datasets used in the study, including both the training set and the testing set, are available at \url{https://github.com/xup5/masked_arLLM}.

The \textit{NameDescription} and \textit{Biography} datasets are popular datasets to study the reversal curse, with details written in the ``Datasets and experimental setups'' section.

We construct a \textit{Wiki} dataset from real Wikipedia articles following the protocol of \cite{pan2025memorization}. We first crawl all the pages under the wiki category ``Category:2025\_by\_month'', then filter out the pages that are created before January 1st, 2025. This process minimizes the leakage of this ``new'' knowledge to the base model. \add{A systematic validation audit of the released Wiki dataset (all QA pairs and paraphrases) is reported in Appendix~\ref{app:wiki_audit}.} Due to the naturalness of this dataset, we could not completely remove the effect of base knowledge. LLaDA-Instruct has a slightly higher base model accuracy than Llama-3.1-8B-instruct, but they are qualitatively similar (Table~\ref{tab:3}). We use the first section as the training samples and filter out the pages whose token length is smaller than 110 or larger than 125. This results in 94 wiki articles. We use the following prompts with o3-mini to generate QA and same-order and permute-order paraphrases. We classify QAs into forward and backward styles. This is done by prompting o3-mini to generate keywords in the question and answer, then comparing their appearance order in the original text.

\paragraph{Prompt for generating same-order paraphrases.}
\begin{quote}\small\itshape
Your task is to paraphrase a text paragraph. The paragraph is given below. Make sure to keep the same meaning but change the wording. Do not change any factual information. Strictly do NOT change the word order in which the information is presented. Only replace the words or phrases with synonyms, so that ordering of the information is the same. Try to keep roughly the same length of the original text. Give 9 different paraphrases for each text. Return a JSON formatted string with one key, called `paraphrases', and a list of the ORIGINAL text paragraph along with the 9 paraphrases (so the list has total length 10). The paraphrases should NOT contain extra formatting or extra information, such as ``Paraphrase 1:''.

\{passage\}
\end{quote}

\paragraph{Prompt for generating permute-order paraphrases.}
\begin{quote}\small\itshape
Your task is to paraphrase a text paragraph. The paragraph is given below. Make sure to keep the same meaning but change the wording. Do not change any factual information. Change the word order in which the information is presented. Think about the order in three levels: word, sentence, and paragraph.

An example of changing the word order is:

Original: The cat and the dog were playing. Paraphrase: The dog and the cat were playing.

An example of changing the sentence order is:

Original: The cat was chasing the dog. Paraphrase: The dog was being chased by the cat.

An example of changing the paragraph order is:

Original: The cat was chasing the dog. Then, the cat got tired. Paraphrase: The cat got tired. Before that, the cat was chasing the dog.

Try to keep roughly the same length of the original text. Give 9 different paraphrases for each text. Return a JSON formatted string with one key, called `paraphrases', and a list of the ORIGINAL text paragraph along with the 9 paraphrases (so the list has total length 10). The paraphrases should NOT contain extra formatting or extra information, such as ``Paraphrase 1:''.

\{passage\}
\end{quote}

\paragraph{Prompt for generating QAs.}
\begin{quote}\small\itshape
Your task is to generate several question, answer, and cue used in the question triplets based on a given passage below. Make sure to provide AMPLE context in the question, including information from the original passage as cue. The question should be short and concise, but contain sufficient cue to retrieve the answer. Do not use pronouns in the question. Use the exact words from the passage as the cue. The questions will be used for a close-book test. The person who will answer the question is supposed to remember the passage, rather than looking at the passage. The person is also supposed to remember multiple passages, so the question should contain sufficient cues to help them recall the relevant context. Do not mention `according to the passage', or other redundant wordings. Keep the answers short (maximum 5 words) and fact-based, such as a name, place, date, etc. Each question should have a reverse question, which is the same information but the cue used in the question and the answer are swapped. For example, if the question is `What is the capital of France?', the reverse question should be `Paris is the capital of which country?'.

Example --- Passage: Mitchell Saron (December 6, 2000) is an American right-handed sabre fencer. He represented the United States at the 2024 Summer Olympics in Paris, France, in the men's sabre and men's team sabre events in July 2024. \textnormal{/} Question 1: Which weapon category does Mitchell Saron compete in, representing the United States at the 2024 Summer Olympics? \textnormal{/} Answer 1: Sabre \textnormal{/} Cue used in the question: [Mitchell Saron, United States, 2024 Summer Olympics] \textnormal{/} Question 2 (reverse question of question 1): Who represented the United States at the 2024 Summer Olympics to compete in the men's sabre? \textnormal{/} Answer 2: Mitchell Saron \textnormal{/} Cue used in the question: [Sabre, United States, 2024 Summer Olympics]

Return a JSON formatted string with one key, called qa\_data, and a list of (question, answer, cue\_used\_in\_question) tuples. Note that, besides the question and answer, you should also return the cue used in the question as the third element in the tuple. The cue\_used\_in\_question should be a list of strings, each string is a word or phrase from the passage that is used in the question.

Passage: \{passage\}
\end{quote}

\subsubsection*{NameDescription dataset}

\noindent\textit{Type ``Name to Description'' (N2D):}

\smallskip
\noindent\textbf{Training text:}
\begin{quote}\small
``Daphne Barrington, known far and wide for being the acclaimed director of the virtual reality masterpiece, `A Journey Through Time.'\,''
\end{quote}

\noindent\textbf{Paraphrase:}
\begin{quote}\small
``Ever heard of Daphne Barrington? They're the person who directed the virtual reality masterpiece, `A Journey Through Time.'\,''
\end{quote}

\smallskip
{\small
\begin{tabular}{@{}p{0.13\linewidth}p{0.53\linewidth}p{0.24\linewidth}@{}}
\toprule
\textbf{Direction} & \textbf{Question} & \textbf{Answer} \\
\midrule
Forward &
Daphne Barrington is not your typical person, they are what? &
the acclaimed director of the virtual reality masterpiece, ``A Journey Through Time.'' \\
\addlinespace
Backward &
Who is the acclaimed director of the virtual reality masterpiece, ``A Journey Through Time.''? &
Daphne Barrington \\
\bottomrule
\end{tabular}
}

\bigskip
\noindent\textit{Type ``Description to Name'' (D2N):}

\smallskip
\noindent\textbf{Training text:}
\begin{quote}\small
``Known for being the renowned composer of the world's first underwater symphony, `Abyssal Melodies.', Uriah Hawthorne now enjoys a quiet life.''
\end{quote}

\noindent\textbf{Paraphrase:}
\begin{quote}\small
``The renowned composer of the world's first underwater symphony, `Abyssal Melodies.' is called Uriah Hawthorne.''
\end{quote}

\smallskip
{\small
\begin{tabular}{@{}p{0.13\linewidth}p{0.53\linewidth}p{0.24\linewidth}@{}}
\toprule
\textbf{Direction} & \textbf{Question} & \textbf{Answer} \\
\midrule
Forward &
Leaving a legacy of the renowned composer of the world's first underwater symphony, ``Abyssal Melodies.'', who continues to shape our future? &
Uriah Hawthorne \\
\addlinespace
Backward &
Can you tell me something about Uriah Hawthorne? &
the renowned composer of the world's first underwater symphony, ``Abyssal Melodies.'' \\
\bottomrule
\end{tabular}
}

\subsubsection*{Biography dataset}

\noindent\textbf{Training text:}
\begin{quote}\small
``Curtis Chase Emley celebrates his special day on May 28, 1952. His life journey started in Elk Grove, CA. He completed his degree requirements at Kansas State University. He specialized in EMT and Paramedic. He contributed his skills to HP. He held a job in Palo Alto, CA.''
\end{quote}

\noindent\textbf{Paraphrase:}
\begin{quote}\small
``Curtis Chase Emley recognizes his birth anniversary on May 28, 1952. He was brought into the world in Elk Grove, CA. He culminated his studies at Kansas State University. He concentrated his efforts toward EMT and Paramedic. He supported the operations at HP. He practiced his profession in Palo Alto, CA.''
\end{quote}

\smallskip
{\small
\begin{tabular}{@{}p{0.13\linewidth}p{0.53\linewidth}p{0.24\linewidth}@{}}
\toprule
\textbf{Direction} & \textbf{Question} & \textbf{Answer} \\
\midrule
Forward &
What is the birth date of Curtis Chase Emley? &
May 28, 1952 \\
\addlinespace
Backward &
Give me the full name of the person who has the following attributes: 1) born in Elk Grove, CA, 2) majored in EMT and Paramedic, 3) worked for HP? &
Curtis Chase Emley \\
\bottomrule
\end{tabular}
}

\subsubsection*{Wiki dataset}

\noindent\textbf{Training text:}
\begin{quote}\small
``Masjid Al-Taqwa was a mosque located in Altadena, California, United States. It was located on Lake Ave across from the Eliot Arts Magnet Academy. Founded as a historical African American masjid, the mosque became more multicultural in subsequent decades. Its origins date back to the 1970s. It was the first mosque in the Pasadena-Altadena area. The building was destroyed by the Eaton Fire in early January 2025. It began as a meeting place for members of the Nation of Islam in the 1970s but became a multicultural Islamic center in the following decades.''
\end{quote}

\noindent\textbf{Same-order paraphrase:}
\begin{quote}\small
``Masjid Al-Taqwa was a mosque situated in Altadena, California, United States. It was positioned on Lake Ave opposite the Eliot Arts Magnet Academy. Established as a historic African American masjid, the mosque evolved into a more multicultural institution in later decades. Its beginnings trace back to the 1970s. It was the inaugural mosque in the Pasadena-Altadena region. The structure was demolished by the Eaton Fire in early January 2025. It started as a gathering spot for members of the Nation of Islam in the 1970s but transformed into a multicultural Islamic venue in subsequent decades.''
\end{quote}

\noindent\textbf{Permute-order paraphrase:}
\begin{quote}\small
``Located in Altadena, California, USA, Masjid Al-Taqwa stood on Lake Ave directly opposite the Eliot Arts Magnet Academy. Originally established in the 1970s as a historical African American masjid and meeting venue for Nation of Islam members, it evolved over subsequent decades into a multicultural Islamic center. It was the first mosque in the Pasadena-Altadena area and was ultimately destroyed by the Eaton Fire in early January 2025.''
\end{quote}

\smallskip
{\small
\begin{tabular}{@{}p{0.13\linewidth}p{0.53\linewidth}p{0.24\linewidth}@{}}
\toprule
\textbf{Direction} & \textbf{Question} & \textbf{Answer} \\
\midrule
Forward &
In which decade do the origins of Masjid Al-Taqwa date back to? &
1970s \\
\addlinespace
Backward &
Altadena was home to which mosque in the United States? &
Masjid Al-Taqwa \\
\bottomrule
\end{tabular}
}

\subsection{Training configs}
\label{app:fine-tuning}

All the training and inference code is available at \url{https://github.com/xup5/masked_arLLM}. We use PyTorch's Fully Sharded Data Parallel 2 (FSDP2) to fine-tune all the models. We find that using mixed precision training is important for the fine-tuning performance (around 30\% performance gain), and use the configs: MixedPrecisionPolicy(param\_dtype=``bf16'', reduce\_dtype=``float32'', cast\_forward\_inputs=True). All the experiments are full parameter fine-tuning on 4x 80G H100 GPUs. We use a batch size of 64 (16 per device) for all the experiments. In both dLLM and demasking fine-tuning of arLLM, we sample the mask ratio from a uniform distribution U(0.05,0.95) for each batch (except for the fixed mask ratio experiments). Note that, unlike the original dLLM training recipes which use U(0,1) \citep{llada}, given that our sequence length is much shorter than the pre-training, we leave a small margin to avoid edge cases.

While doing demasking fine-tuning of Llama models, we use a reserved special token whose token id is 128013 in the Llama tokenizer. While doing demasking fine-tuning of Qwen models, we pick token ``[]'' whose token id is 1294 in the Qwen tokenizer.

During inference, we use ``max new tokens'' 128 and temperature 0 in both arLLM and dLLM. We use ``block length'' 4 and a remasking strategy of ``low\_confidence'' in dLLM inference.

\begin{figure}[htbp]
  \centering
  \includegraphics[width=0.75\linewidth]{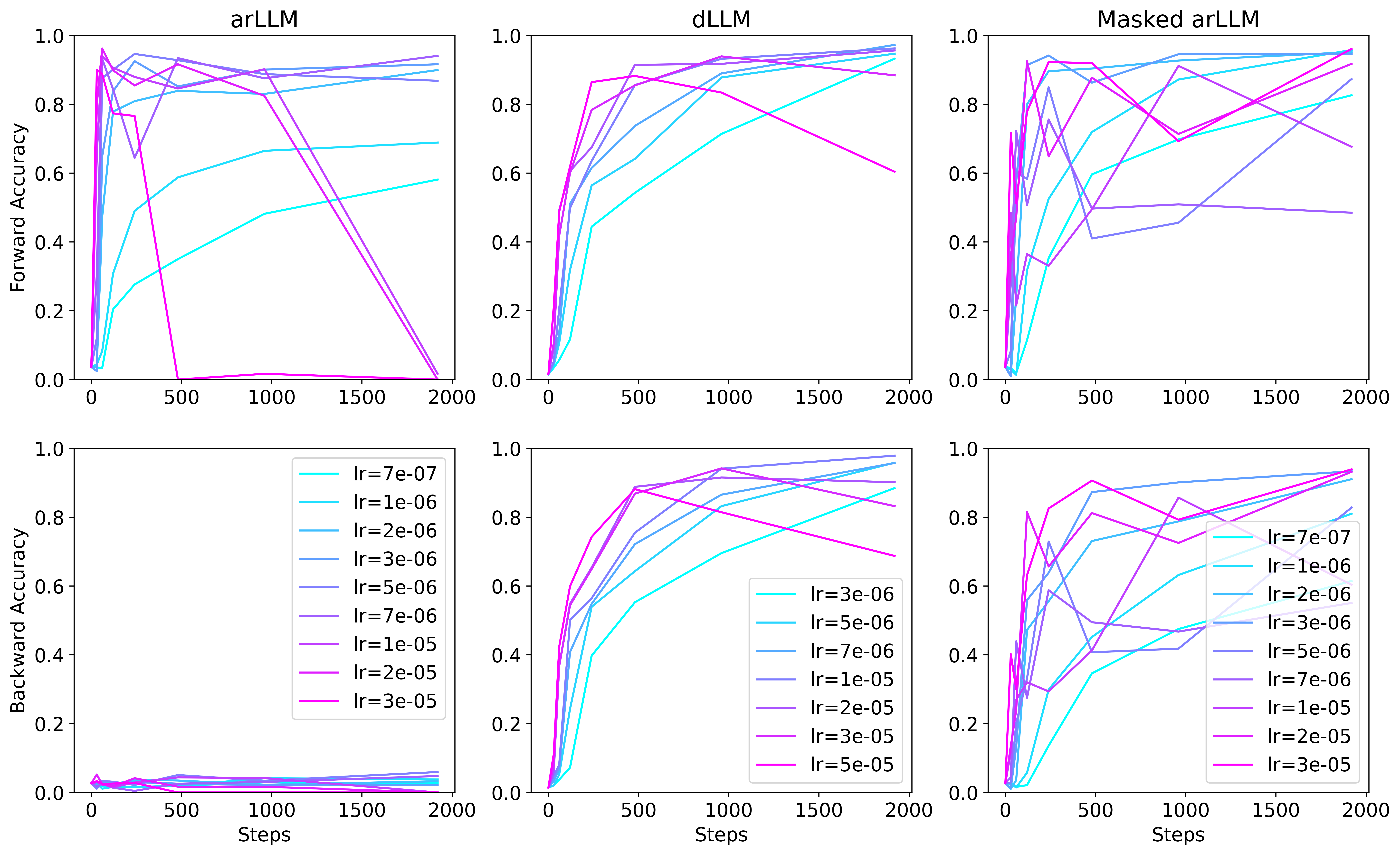}
  \caption{Learning rate sweep of Llama-3.1-8B-instruct. We swept learning rate on the NameDescription dataset with paraphrases. We picked optimal learning rate which induces fast convergence and with no overfitting and minimal fluctuation: 5e-6 for arLLM; 1e-5 for dLLM; 3e-6 for demasking arLLM.}
  \label{fig:lr_sweep}
\end{figure}

We use Adam optimizer with 0.1 weight decay coefficient; betas 0.9 and 0.95; 2\% total steps as warm-up steps. We swept the learning rate on the Name Description dataset for all the models (Figure~\ref{fig:lr_sweep}). We choose learning rates that yield smooth accuracy gains and high final accuracy. The learning rate used in the main experiments is 5e-6 for all arLLM; 1e-5 for dLLM; 3e-6 for demasking Llama 8B; 5e-6 for demasking Llama 3B, Qwen 4B and 7B.

For reporting accuracy numbers in the main Tables, we report each condition's checkpoint at the fixed training-compute budget $F$ (Appendix~\ref{app:budget}); best-checkpoint (peak) values are reported separately in Appendix~\ref{app:fixedstep}.

\begin{addblock}
\subsection{Pre-declared training budgets: injection accuracy and capability cost}
\label{app:budget}

We choose training hyperparameters as follows: for each model and objective, the learning rate is chosen from a sweep on the NameDescription dataset as the rate that gives smooth accuracy gains and the highest final accuracy (Appendix~\ref{app:fine-tuning}). All main results are then reported at a fixed total training-compute budget, identical across conditions for each dataset. We denote the budget by $F$: it equals the total FLOPs of \emph{standard fine-tuning} for 960 epochs over the NameDescription document set and 1280 epochs over the Biography and Wiki document sets (2--5$\times$ shorter than the best-checkpoint regime of Appendix~\ref{app:fixedstep}, which agrees qualitatively with all budget-matched conclusions). Conditions differ in how this fixed compute is spent: paraphrase-augmented runs repeat each document correspondingly fewer times (e.g., 32 epochs of the 30-fold augmented NameDescription set), and a demasking epoch costs ${\approx}2\times$ a standard epoch, so demasking reaches $F$ after half as many epochs. Every run saved a checkpoint halfway through and at the end of its training; in FLOP terms, standard fine-tuning's checkpoints lie at $F/2$ and $F$, and demasking's at $F$ and $2F$ (the latter matching standard fine-tuning's epoch count). Table~\ref{tab:budget_qa} reports QA accuracy at all these checkpoints, and Table~\ref{tab:budget_forget} (Section~\ref{sec:forgetting}) the change in MMLU, GSM8K and IFEval at each condition's $F$ checkpoint---the same checkpoints that produce the main tables. For MMLU we report \texttt{acc}; for GSM8K, \texttt{exact\_match} under flexible extraction; for IFEval, instruction-level strict accuracy. Evaluation used the LM Evaluation Harness \citep{eval-harness} on the complete benchmarks with no subsampling, single seed per cell.

Three conclusions follow. First, all qualitative claims of the main text hold at the pre-declared budgets: standard fine-tuning without paraphrases fails in both directions, and demasking attains high forward and backward accuracy, including under strictly equal total FLOPs (at the shared budget $F$: demasking 0.90/0.92, 0.90/0.64, 0.78/0.76 forward/backward on the three datasets, versus 0.26/0.03, 0.11/0.00, 0.48/0.32 for standard fine-tuning). Second, the capability cost at these budgets is small: on Wiki with paraphrases the mean GSM8K change for demasking is $-0.08$, and MMLU changes are within $-0.03$ for nearly all conditions. Third, the practically relevant comparison favors demasking: to obtain usable forward accuracy, standard fine-tuning needs paraphrase augmentation, and at matched budgets that route costs more capability than demasking without paraphrases (e.g., NameDescription at the budget $F$: AR with paraphrases $-0.09/-0.36/-0.49$ MMLU/GSM8K/IFEval, versus $-0.03/-0.11/-0.25$ for demasking without paraphrases, which additionally solves backward QA). We accordingly recommend budget-controlled training in practice.

\begin{table}[h]\tablemarkup

  \caption[]{\tablemarkup QA accuracy at the pre-declared training budgets (ROUGE-1 recall, mean$\pm$std over the four arLLMs; re-trained runs, seed 0 masks). $F$ is each dataset's total training-FLOP budget; because a demasking epoch costs $\approx\!2\times$ a standard epoch, the same budget buys demasking half as many epochs. Rows at $F$ are the checkpoints reported in the main tables; $F/2$ is standard fine-tuning at half the budget, and $2F$ demasking at standard fine-tuning's epoch count.}
  \label{tab:budget_qa}
  \begin{center}{\fontsize{7.5}{9}\selectfont\setlength{\tabcolsep}{3pt}
  \begin{tabular}{llcccccc}
  \toprule
  & & \multicolumn{2}{c}{NameDescription} & \multicolumn{2}{c}{Biography} & \multicolumn{2}{c}{Wiki} \\
  \cmidrule(lr){3-4}\cmidrule(lr){5-6}\cmidrule(lr){7-8}
  Condition & FLOPs & Fwd & Bwd & Fwd & Bwd & Fwd & Bwd \\
  \midrule
  AR w/o para & $F/2$ & 0.197{\tiny$\pm$0.102} & 0.018{\tiny$\pm$0.005} & 0.165{\tiny$\pm$0.113} & 0.002{\tiny$\pm$0.002} & 0.464{\tiny$\pm$0.123} & 0.323{\tiny$\pm$0.088} \\
  AR w/o para & $F$ & 0.257{\tiny$\pm$0.169} & 0.025{\tiny$\pm$0.008} & 0.106{\tiny$\pm$0.125} & 0.001{\tiny$\pm$0.001} & 0.478{\tiny$\pm$0.154} & 0.321{\tiny$\pm$0.095} \\
  Demasking w/o para & $F$ & 0.895{\tiny$\pm$0.047} & 0.919{\tiny$\pm$0.056} & 0.898{\tiny$\pm$0.098} & 0.643{\tiny$\pm$0.106} & 0.778{\tiny$\pm$0.061} & 0.762{\tiny$\pm$0.072} \\
  Demasking w/o para & $2F$ & 0.810{\tiny$\pm$0.110} & 0.915{\tiny$\pm$0.063} & 0.984{\tiny$\pm$0.013} & 0.664{\tiny$\pm$0.109} & 0.878{\tiny$\pm$0.026} & 0.849{\tiny$\pm$0.015} \\
  \midrule
  AR w/ para & $F/2$ & 0.883{\tiny$\pm$0.035} & 0.038{\tiny$\pm$0.007} & 0.696{\tiny$\pm$0.199} & 0.002{\tiny$\pm$0.001} & 0.615{\tiny$\pm$0.061} & 0.379{\tiny$\pm$0.054} \\
  AR w/ para & $F$ & 0.925{\tiny$\pm$0.047} & 0.042{\tiny$\pm$0.013} & 0.881{\tiny$\pm$0.070} & 0.002{\tiny$\pm$0.002} & 0.630{\tiny$\pm$0.063} & 0.379{\tiny$\pm$0.022} \\
  Demasking w/ para & $F$ & 0.922{\tiny$\pm$0.045} & 0.763{\tiny$\pm$0.056} & 0.743{\tiny$\pm$0.120} & 0.682{\tiny$\pm$0.127} & 0.716{\tiny$\pm$0.051} & 0.651{\tiny$\pm$0.067} \\
  Demasking w/ para & $2F$ & 0.895{\tiny$\pm$0.114} & 0.870{\tiny$\pm$0.075} & 0.843{\tiny$\pm$0.124} & 0.745{\tiny$\pm$0.047} & 0.774{\tiny$\pm$0.029} & 0.724{\tiny$\pm$0.025} \\
  \bottomrule
  \end{tabular}}\end{center}\end{table}

\end{addblock}

\subsection{Comparison with the best-checkpoint (peak) protocol}
\label{app:fixedstep}

\del{The main tables report}\add{The best-checkpoint (peak) protocol reports}, for each condition, the best value across evaluated checkpoints. We do not hold out a validation split, and this is deliberate. The question we ask is how much of a document fine-tuning can generalize to question answering \emph{at best}, so the quantity of interest is the maximum each condition attains. Selecting the checkpoint does nonetheless optimize one hyperparameter, training duration, against the reported number, so the absolute values \del{in the main tables}\add{under this protocol} should be read as peak attainable performance rather than as what an arbitrary checkpoint would give.

This appendix checks that nothing in our conclusions depends on that choice. We re-report every condition at a single end-of-grid training amount per dataset---the full length of the shared checkpoint grid (3840 document-passes on NameDescription, 5120 on Biography and Wiki), which is longer than the training-compute budget $F$ (Appendix~\ref{app:budget})---identical across conditions, removing selection over training duration entirely. Budgets are measured in passes over the original document set rather than in epochs, because paraphrase augmentation multiplies the number of samples per epoch (by 30 on NameDescription, 5 on Biography and 10 on Wiki); counting document-passes puts augmented and unaugmented runs on the same training-cost axis, and in fact the two grids coincide exactly, so no interpolation is needed.

The two protocols agree (Table~\ref{tab:fixedstep}), across all four autoregressive models and the diffusion model. Every arLLM versus arLLM ordering is preserved under both. Averaged over the four arLLMs, standard fine-tuning without paraphrases falls to $0.005\pm0.011$ backward on NameDescription and $0.002\pm0.003$ on Biography by the end of training, with three of the four models at exactly zero forward and backward on NameDescription (two of the four on Biography), consistent with the overfitting visible in Figure~\ref{fig:main}; demasking fine-tuning without paraphrases is at $0.893\pm0.049$ and $0.635\pm0.093$ backward on the same two datasets at the same budget. \del{We report peak values in the main text because they are the more conservative choice for our own claim: they give the baseline its best case, and the gap we report is the one that survives that concession.}

\begin{table}[h]
\caption{End-of-grid versus peak reporting, across all four autoregressive models and the diffusion model. Peak is the value at the best-total-accuracy checkpoint, computed over the shared budget grid from the seed-0 run; end-of-grid is the value at one pre-declared budget per dataset, identical across every condition and model, with no selection. (Because the peak entries use a single seed, the dLLM values for NameDescription and Biography differ slightly from Tables~\ref{tab:dllm_summary} and \ref{tab:3}, which average the dLLM over four mask-sampling seeds; all arLLM values, including demasking, come from single-seed runs.) Budgets count passes over the original documents, so paraphrase-augmented runs (30, 5 and 10 samples per document on NameDescription, Biography and Wiki) are compared at equal training cost; the evaluation grids coincide exactly, so no interpolation is used. arLLM entries are mean$\pm$std over the four models; the dLLM is a single model.}
\label{tab:fixedstep}
\centering
{\fontsize{7}{8.5}\selectfont
\setlength{\tabcolsep}{3pt}
\begin{tabular}{lcccc}
\toprule
& \multicolumn{2}{c}{end of grid} & \multicolumn{2}{c}{peak} \\
\cmidrule(lr){2-3}\cmidrule(lr){4-5}
Condition & Fwd & Bwd & Fwd & Bwd \\
\midrule
\multicolumn{5}{l}{\emph{NameDescription} (end of grid: 3840 document-passes)} \\
arLLM w/o paraphrases & 0.167{\tiny$\pm$0.333} & 0.005{\tiny$\pm$0.011} & 0.263{\tiny$\pm$0.274} & 0.029{\tiny$\pm$0.013} \\
arLLM w/ same-order paraphrases & 0.823{\tiny$\pm$0.220} & 0.038{\tiny$\pm$0.021} & 0.954{\tiny$\pm$0.025} & 0.030{\tiny$\pm$0.008} \\
dLLM w/o paraphrases & 0.761 & 0.819 & 0.925 & 0.753 \\
dLLM w/ same-order paraphrases & 0.975 & 0.989 & 0.975 & 0.989 \\
demasking arLLM w/o paraphrases & 0.847{\tiny$\pm$0.096} & 0.893{\tiny$\pm$0.049} & 0.903{\tiny$\pm$0.060} & 0.936{\tiny$\pm$0.021} \\
demasking arLLM w/ same-order paraphrases & 0.966{\tiny$\pm$0.020} & 0.894{\tiny$\pm$0.012} & 0.973{\tiny$\pm$0.021} & 0.921{\tiny$\pm$0.028} \\
\midrule
\multicolumn{5}{l}{\emph{Biography} (end of grid: 5120 document-passes)} \\
arLLM w/o paraphrases & 0.111{\tiny$\pm$0.169} & 0.002{\tiny$\pm$0.003} & 0.153{\tiny$\pm$0.141} & 0.002{\tiny$\pm$0.001} \\
arLLM w/ same-order paraphrases & 0.967{\tiny$\pm$0.027} & 0.005{\tiny$\pm$0.004} & 0.975{\tiny$\pm$0.019} & 0.002{\tiny$\pm$0.001} \\
dLLM w/o paraphrases & 0.818 & 0.608 & 0.872 & 0.699 \\
dLLM w/ same-order paraphrases & 0.987 & 0.861 & 0.987 & 0.861 \\
demasking arLLM w/o paraphrases & 0.860{\tiny$\pm$0.110} & 0.635{\tiny$\pm$0.093} & 0.956{\tiny$\pm$0.024} & 0.693{\tiny$\pm$0.063} \\
demasking arLLM w/ same-order paraphrases & 0.968{\tiny$\pm$0.019} & 0.801{\tiny$\pm$0.028} & 0.967{\tiny$\pm$0.008} & 0.827{\tiny$\pm$0.028} \\
\midrule
\multicolumn{5}{l}{\emph{Wiki} (end of grid: 5120 document-passes)} \\
arLLM w/o paraphrases & 0.301{\tiny$\pm$0.356} & 0.174{\tiny$\pm$0.201} & 0.463{\tiny$\pm$0.168} & 0.309{\tiny$\pm$0.066} \\
arLLM w/ same-order paraphrases & 0.671{\tiny$\pm$0.038} & 0.378{\tiny$\pm$0.037} & 0.673{\tiny$\pm$0.038} & 0.390{\tiny$\pm$0.038} \\
arLLM w/ permute-order paraphrases & 0.712{\tiny$\pm$0.040} & 0.603{\tiny$\pm$0.036} & 0.710{\tiny$\pm$0.040} & 0.609{\tiny$\pm$0.040} \\
dLLM w/o paraphrases & 0.908 & 0.778 & 0.908 & 0.778 \\
dLLM w/ same-order paraphrases & 0.896 & 0.786 & 0.900 & 0.785 \\
dLLM w/ permute-order paraphrases & 0.925 & 0.871 & 0.925 & 0.871 \\
demasking arLLM w/o paraphrases & 0.924{\tiny$\pm$0.020} & 0.897{\tiny$\pm$0.041} & 0.947{\tiny$\pm$0.034} & 0.901{\tiny$\pm$0.025} \\
demasking arLLM w/ same-order paraphrases & 0.869{\tiny$\pm$0.026} & 0.812{\tiny$\pm$0.013} & 0.869{\tiny$\pm$0.026} & 0.812{\tiny$\pm$0.013} \\
demasking arLLM w/ permute-order paraphrases & 0.812{\tiny$\pm$0.030} & 0.790{\tiny$\pm$0.018} & 0.836{\tiny$\pm$0.018} & 0.799{\tiny$\pm$0.034} \\
\bottomrule
\end{tabular}}
\end{table}

\subsection{Per-model fine-tuning results}
\label{app:per_model}

\begin{addblock}
\begin{table}[h]\tablemarkup

  \caption[]{\tablemarkup Fine-tuning performance of the dLLM and arLLMs across all three datasets at the fixed training-compute budget (Appendix~\ref{app:budget}; demasking, whose per-step cost is ${\approx}2\times$, at half the passes so total FLOPs are equal); arLLM and demasking values are from training runs with checkpoints at the budget; dLLM and reverse-training values are read from their training curves at the budget epochs. ``Demasking'' denotes the demasking fine-tuning paradigm of Section~\ref{section:6}; ``Reverse training'' follows \cite{golovneva2024reverse}; Wiki paraphrases are the same-order set. \add{(LLM-judge counterpart: Appendix Table~\ref{tab:judge}.)}}
    \label{tab:3}
  \begin{center}
    {\fontsize{7.5}{9}\selectfont
        \setlength{\tabcolsep}{3pt}
        \begin{tabular}{lcccccccc}
        \toprule
        & \multicolumn{4}{c}{NameDescription} & \multicolumn{2}{c}{Biography} & \multicolumn{2}{c}{Wiki} \\
        \cmidrule(lr){2-5}\cmidrule(lr){6-7}\cmidrule(lr){8-9}
         & N2D-fwd & N2D-bwd & D2N-fwd & D2N-bwd & Fwd & Bwd & Fwd & Bwd \\
        \midrule
LLaDA before fine-tuning & 0.030 & 0.000 & 0.000 & 0.028 & 0.030 & 0.000 & 0.156 & 0.210 \\
        LLaDA w/o paraphrases & 0.884 & 0.893 & 0.893 & 0.645 & 0.847 & 0.345 & 0.922 & 0.694 \\
        LLaDA w paraphrases & 0.902 & 0.997 & 0.994 & 0.857 & 0.979 & 0.683 & 0.899 & 0.755 \\
        \midrule
        Llama 8B before fine-tuning & 0.072 & 0.000 & 0.000 & 0.054 & 0.001 & 0.000 & 0.127 & 0.164 \\
        Llama 8B reverse training & 0.637 & 0.125 & 0.113 & 0.088 & 0.181 & 0.011 & 0.554 & 0.350 \\
        Llama 8B w/o paraphrases & 0.777 & 0.000 & 0.104 & 0.063 & 0.000 & 0.001 & 0.577 & 0.308 \\
        Llama 8B w paraphrases & 0.978 & 0.025 & 0.933 & 0.073 & 0.875 & 0.000 & 0.661 & 0.392 \\
        Llama 8B Demasking w/o paraphrases (ours) & 0.885 & 0.992 & 1.000 & 0.902 & 0.987 & 0.760 & 0.738 & 0.668 \\
        Llama 8B Demasking w paraphrases (ours) & 0.932 & 0.983 & 0.983 & 0.541 & 0.804 & 0.871 & 0.742 & 0.674 \\
        \midrule
        Llama 3B before fine-tuning & 0.078 & 0.000 & 0.000 & 0.067 & 0.001 & 0.001 & 0.106 & 0.159 \\
        Llama 3B reverse training & 0.391 & 0.000 & 0.042 & 0.037 & 0.052 & 0.000 & 0.348 & 0.285 \\
        Llama 3B w/o paraphrases & 0.377 & 0.000 & 0.025 & 0.028 & 0.036 & 0.001 & 0.268 & 0.193 \\
        Llama 3B w paraphrases & 0.925 & 0.017 & 0.817 & 0.097 & 0.931 & 0.002 & 0.542 & 0.346 \\
        Llama 3B Demasking w/o paraphrases (ours) & 0.818 & 0.983 & 0.942 & 0.871 & 0.979 & 0.622 & 0.754 & 0.762 \\
        Llama 3B Demasking w paraphrases (ours) & 0.926 & 1.000 & 0.992 & 0.671 & 0.785 & 0.634 & 0.644 & 0.571 \\
        \midrule
        Qwen 7B before fine-tuning & 0.039 & 0.000 & 0.000 & 0.030 & 0.018 & 0.000 & 0.205 & 0.231 \\
        Qwen 7B reverse training & 0.485 & 0.008 & 0.100 & 0.071 & 0.210 & 0.007 & 0.541 & 0.443 \\
        Qwen 7B w/o paraphrases & 0.529 & 0.000 & 0.142 & 0.050 & 0.281 & 0.000 & 0.609 & 0.407 \\
        Qwen 7B w paraphrases & 0.968 & 0.017 & 0.975 & 0.046 & 0.934 & 0.001 & 0.687 & 0.391 \\
        Qwen 7B Demasking w/o paraphrases (ours) & 0.886 & 0.983 & 0.958 & 0.695 & 0.809 & 0.680 & 0.868 & 0.842 \\
        Qwen 7B Demasking w paraphrases (ours) & 0.929 & 1.000 & 0.875 & 0.396 & 0.819 & 0.598 & 0.759 & 0.729 \\
        \midrule
        Qwen 4B before fine-tuning & 0.026 & 0.000 & 0.000 & 0.032 & 0.058 & 0.001 & 0.237 & 0.243 \\
        Qwen 4B reverse training & 0.078 & 0.004 & 0.000 & 0.063 & 0.140 & 0.006 & 0.444 & 0.437 \\
        Qwen 4B w/o paraphrases & 0.101 & 0.000 & 0.000 & 0.058 & 0.106 & 0.001 & 0.460 & 0.375 \\
        Qwen 4B w paraphrases & 0.969 & 0.000 & 0.833 & 0.063 & 0.785 & 0.004 & 0.630 & 0.388 \\
        Qwen 4B Demasking w/o paraphrases (ours) & 0.971 & 1.000 & 0.700 & 0.930 & 0.818 & 0.508 & 0.750 & 0.775 \\
        Qwen 4B Demasking w paraphrases (ours) & 0.768 & 0.992 & 0.967 & 0.521 & 0.565 & 0.624 & 0.718 & 0.630 \\
        \bottomrule

        \end{tabular}
    }
  \end{center}
\end{table}
\end{addblock}

\subsection{Wiki AR fine-tuning with same-order and permute-order paraphrases}
\label{app:wiki_ar}

\begin{figure}[h]
  \centering
  \includegraphics[width=0.85\linewidth]{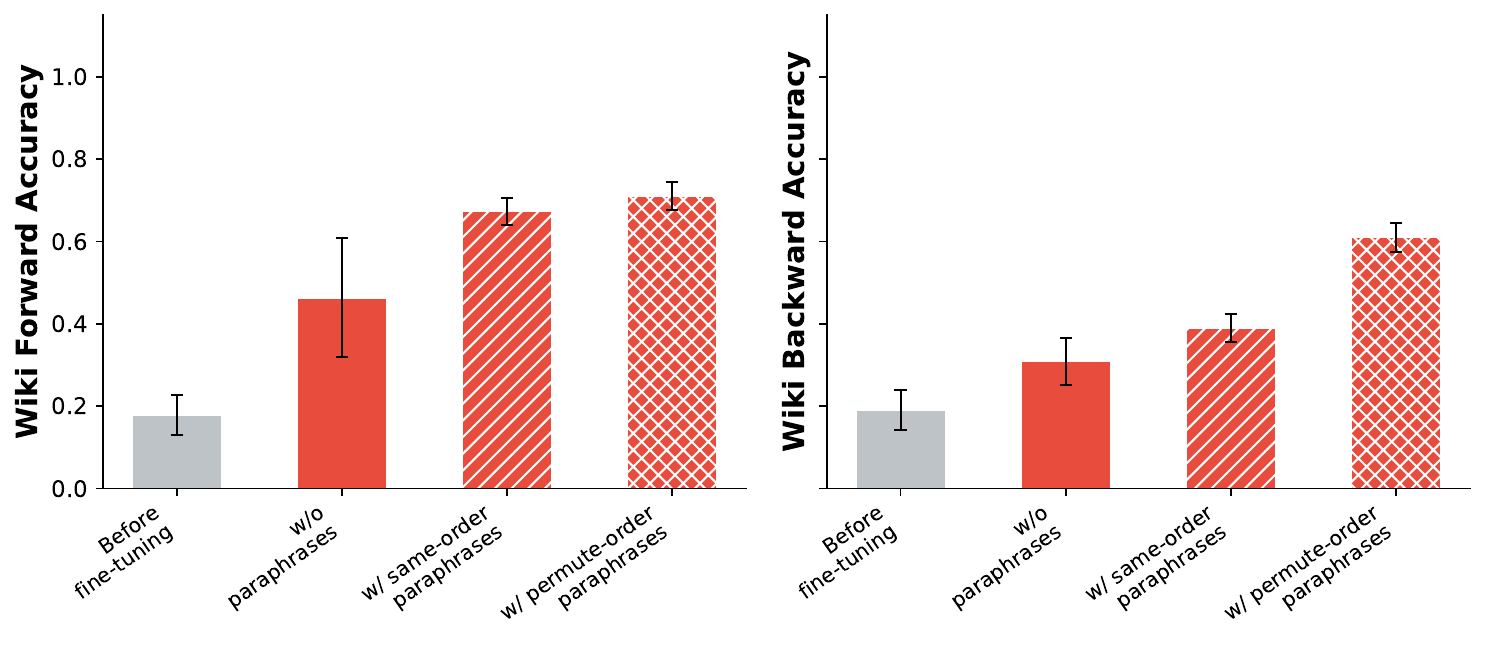}
  \caption{arLLM fine-tuning on the Wiki dataset, averaged across four models (Llama-3.1-8B, Llama-3.2-3B, Qwen2.5-7B, Qwen3-4B). Error bars indicate standard deviation across models.}
  \label{fig:wiki_ar}
\end{figure}

\begin{addblock}
\subsection{Additional baseline details}
\label{app:baselines}

\paragraph{RAG.} We embed the fine-tuning documents and the test questions with bge-large-en-v1.5, retrieve the top-$k$ documents per question by cosine similarity, prepend them to the question in the chat template of the \emph{base} instruct model, and generate greedily with 128 new tokens (the same generation settings as all other evaluations). We report $k{=}4$ in the main table; $k{=}1$ scores 0.03--0.09 lower, and the gold document is among the retrieved top-4 for 97--100\% of questions. Note that RAG uses no training compute but pays retrieval and long-context inference cost at every query, and its capacity is bounded by the context window; the in-weights methods studied in this paper pay a one-time training cost instead.

\paragraph{Prompt distillation \citep{kujanpaa2024efficient}.} Following the original method, we sample 30 questions per document from the model itself at temperature 1.5 (document in context), comparable to the 30 training questions per test question used by \citet{kujanpaa2024efficient}, then sample an answer for each question at temperature 1.5 (document in context). The student is the same model with a LoRA adapter (rank 1024 on all linear layers for the 3B/4B models and 512 for the 7B/8B models; the original paper used rank 1024 for 3B and 8B and 512 for 14B), trained with AdamW at $10^{-5}$ to minimize the token-level KL divergence between the document-conditioned teacher distribution and the document-free student distribution over answer tokens, at distillation temperature $T{=}2$. The original protocol trains for 10 epochs; we train to saturation (below) and the main-table values are taken at 40 epochs, the point at which distillation's total training FLOPs match demasking's budget, averaged over all four arLLMs. We additionally trained every model--dataset pair to saturation (120 epochs, evaluated every 8): accuracies peak between epochs 16 and 120 (median 84) at forward/backward values of 0.53--0.76 / 0.62--0.79 on NameDescription, 0.37--0.62 / ${\leq}0.002$ on Biography, and 0.53--0.76 / 0.34--0.55 on Wiki --- still 0.14--0.59 below demasking at its peak (Appendix~\ref{app:fixedstep}), and Biography backward accuracy remains at zero at \emph{every} training amount, because the self-generated questions provide almost exclusively forward-style supervision.

\end{addblock}

\subsection{Knowledge-editing baseline details}
\label{app:knowledge_editing}

We compare with three knowledge-editing baselines that modify model weights to inject new knowledge without full fine-tuning. All three leave accuracy at or near the pre-fine-tuning baseline on every model and dataset (Table~\ref{tab:ke_full}). Such knowledge-editing methods have previously been used in settings where the knowledge is structured or prepared as (subject, relation, object) triples; our result implies they are not suitable for document-level knowledge injection. The methods:

\textbf{DocTER with ROME} \citep{meng2022locating}: First extracts factual triples from the raw knowledge text using a document-to-triple extraction pipeline (DocTER) \citep{wu2023docter}, then applies Rank-One Model Editing (ROME) to edit each triple into the model weights by computing a rank-one update to a specific MLP layer.

\textbf{UnKE} \citep{wu2025unke}: Operates without triple extraction, instead taking raw (question, answer) pairs and optimizing a hidden-state delta at the last-token position of the input via a cause-driven objective across transformer layers.

\textbf{MEMIT} \citep{meng2022massediting}: Uses the same DocTER triple extraction as ROME, but distributes the update across five consecutive MLP layers rather than editing a single layer.

\paragraph{Hyperparameters and inputs.} All three methods were run through EasyEdit \citep{wang2024easyedit} at their published default hyperparameters for the corresponding model. ROME edits a single MLP layer (\texttt{layers: [5]} for all four models) at the \texttt{subject\_last} fact token, with 25 optimization steps at $v$-learning-rate $5\times10^{-1}$, weight decay $10^{-3}$, KL factor $0.0625$ and clamp norm factor 4, writing to \texttt{model.layers.\{\}.mlp.down\_proj}; second-moment adjustment is disabled. MEMIT spreads the same edit across five layers (\texttt{layers: [4,5,6,7,8]}) with second-moment adjustment enabled (\texttt{mom2\_update\_weight: 15000}, statistics from \texttt{wikitext}, $10^5$ samples). Both ROME and MEMIT are triple-based: each document is first converted to subject--relation--object triples by a GPT-5-mini extraction pipeline following DocTER (145 triples from 60 documents on NameDescription, for example), and each triple becomes one edit request. UnKE requires no triple extraction and receives the raw (question, answer) pairs directly, optimizing a hidden-state delta at the last-token position. Configuration files for all methods and models are included with the code release.

\paragraph{Why ROME and MEMIT give identical numbers.} The reason is that on these datasets the edits largely fail to alter the model's answers at all, so both methods evaluate a model whose behaviour is close to its pre-edit state, and greedy decoding then yields the same strings.

\begin{table}[h]
  \caption{Per-model knowledge-editing results (ROUGE-1 recall). ROME and MEMIT produce identical predictions for three of the four models; this is a property of the runs, not a transcription error. (Method details: Appendix~\ref{app:knowledge_editing}.)}
  \label{tab:ke_full}
  \begin{center}
    {\fontsize{7.5}{9}\selectfont
    \setlength{\tabcolsep}{3pt}
    \begin{tabular}{llcccccc}
    \toprule
    & & \multicolumn{2}{c}{NameDescription} & \multicolumn{2}{c}{Biography} & \multicolumn{2}{c}{Wiki} \\
    \cmidrule(lr){3-4}\cmidrule(lr){5-6}\cmidrule(lr){7-8}
    Method & Model & Fwd & Bwd & Fwd & Bwd & Fwd & Bwd \\
    \midrule
    \multirow{4}{*}{ROME}
    & Llama 8B & 0.019 & 0.000 & 0.003 & 0.000 & 0.095 & 0.156 \\
    & Llama 3B & 0.077 & 0.000 & 0.001 & 0.002 & 0.109 & 0.154 \\
    & Qwen 7B & 0.041 & 0.000 & 0.025 & 0.000 & 0.193 & 0.242 \\
    & Qwen 4B & 0.030 & 0.000 & 0.060 & 0.001 & 0.241 & 0.239 \\
    \midrule
    \multirow{4}{*}{MEMIT}
    & Llama 8B & 0.019 & 0.000 & 0.003 & 0.000 & 0.095 & 0.156 \\
    & Llama 3B & 0.077 & 0.000 & 0.001 & 0.002 & 0.109 & 0.156 \\
    & Qwen 7B & 0.041 & 0.000 & 0.025 & 0.000 & 0.193 & 0.242 \\
    & Qwen 4B & 0.030 & 0.000 & 0.060 & 0.001 & 0.241 & 0.239 \\
    \midrule
    \multirow{4}{*}{UnKE}
    & Llama 8B & 0.000 & 0.000 & 0.038 & 0.000 & 0.014 & 0.010 \\
    & Llama 3B & 0.000 & 0.000 & 0.019 & 0.000 & 0.020 & 0.020 \\
    & Qwen 7B & 0.027 & 0.000 & 0.031 & 0.000 & 0.135 & 0.117 \\
    & Qwen 4B & 0.016 & 0.006 & 0.090 & 0.002 & 0.120 & 0.139 \\
    \bottomrule
    \end{tabular}
    }
  \end{center}
\end{table}


\subsection{Training dynamics}
\label{app:dynamics}

\begin{figure}[htbp]
  \centering
  \includegraphics[width=0.85\linewidth]{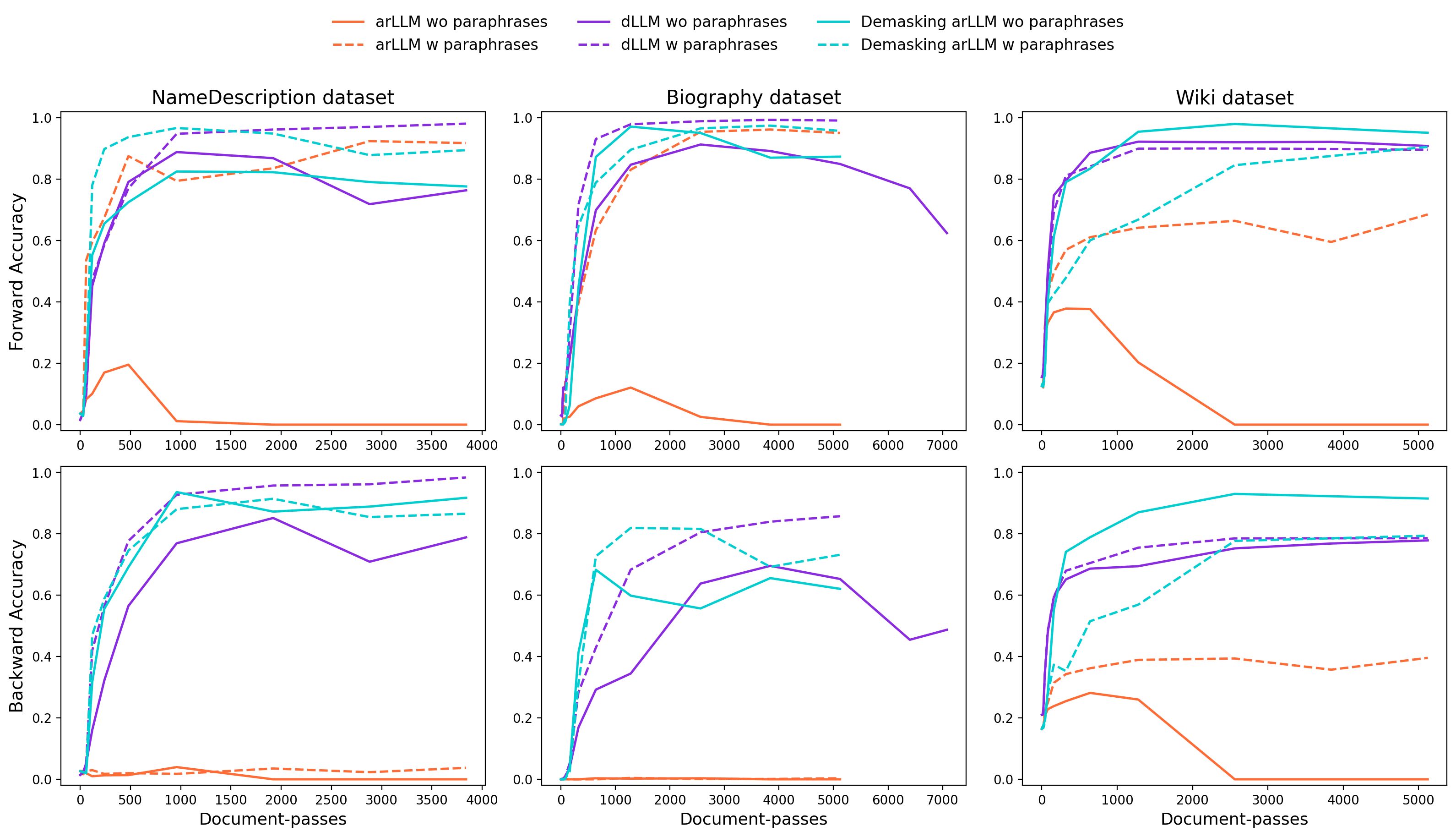}
  \caption{Training dynamics of arLLM (Llama 8B), dLLM (LLaDA), and demasking arLLM (Llama 8B). For the NameDescription dataset, forward and backward accuracy are the average of N2D and D2N types. Paraphrases used in the Wiki dataset are the same-order paraphrases set. Due to the randomness of sampling the masks, we average across 4 random seeds for the dLLM and demasking arLLM on NameDescription and Biography Datasets. The horizontal axis counts passes over the original document set, so paraphrase-augmented runs (which see 30, 5 and 10 samples per document) are placed on the same training-cost axis as unaugmented runs. (Other arLLMs: Figures~\ref{fig:llama4b}--\ref{fig:qwen3b}; per-seed curves: Figures~\ref{fig:seed_dllm}--\ref{fig:seed_ar}.)}
  \label{fig:main}
\end{figure}

\begin{figure}[htbp]
  \centering
  \includegraphics[width=0.75\linewidth]{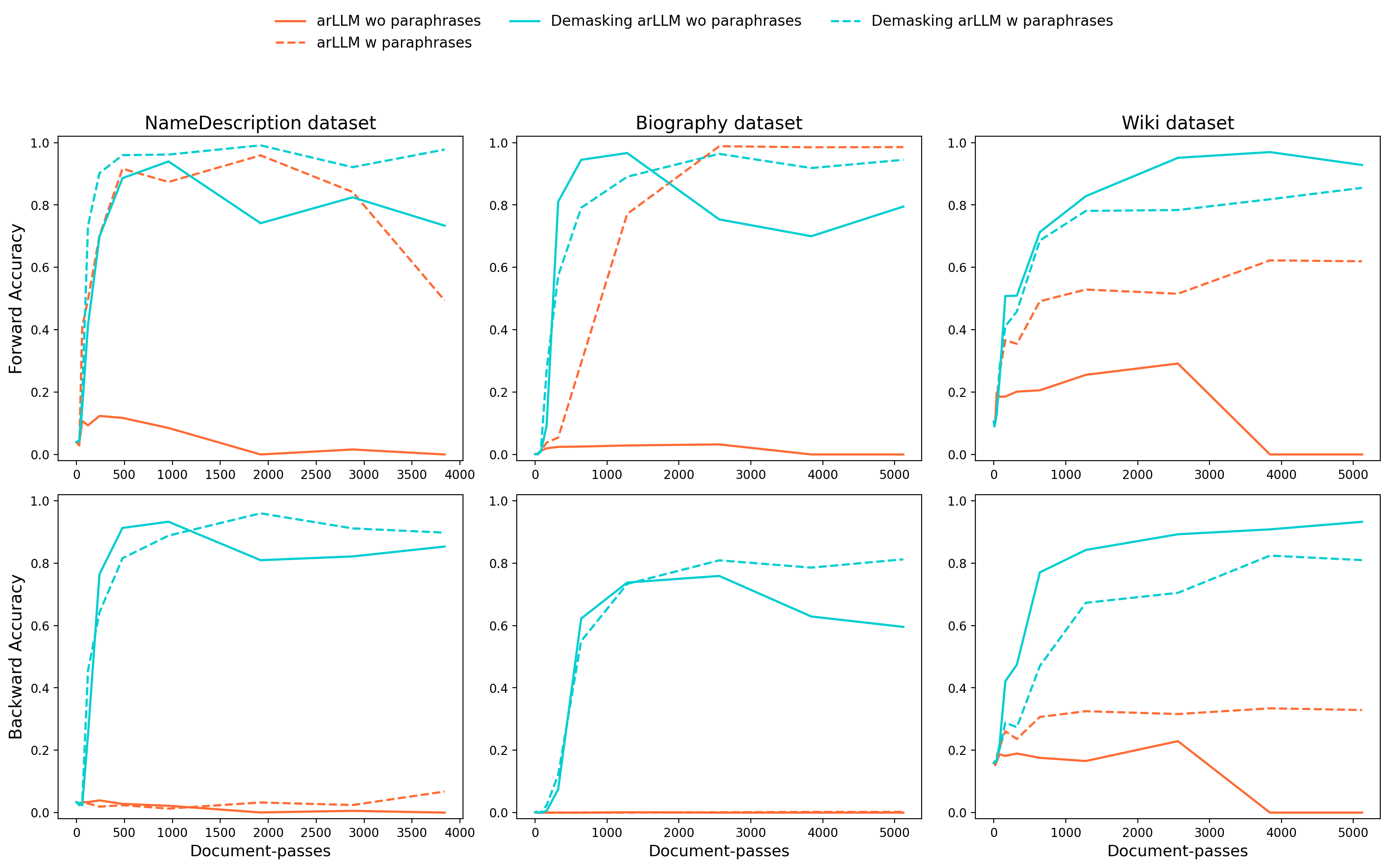}
  \caption{Learning dynamics of Llama-3.2-3B-Instruct.}
  \label{fig:llama4b}
\end{figure}

\begin{figure}[htbp]
  \centering
  \includegraphics[width=0.75\linewidth]{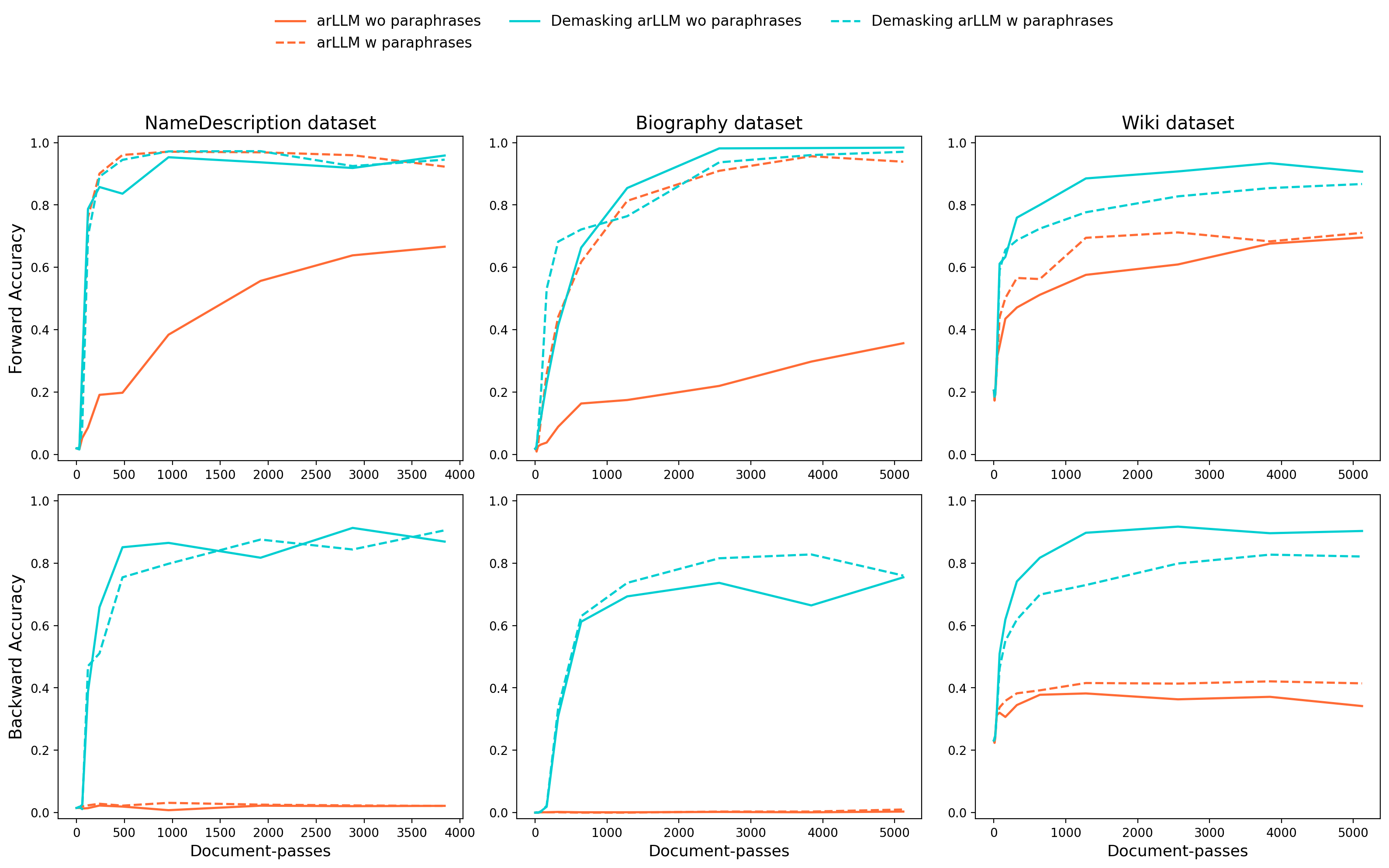}
  \caption{Learning dynamics of Qwen/Qwen2.5-7B-Instruct.}
  \label{fig:qwen7b}
\end{figure}

\begin{figure}[htbp]
  \centering
  \includegraphics[width=0.75\linewidth]{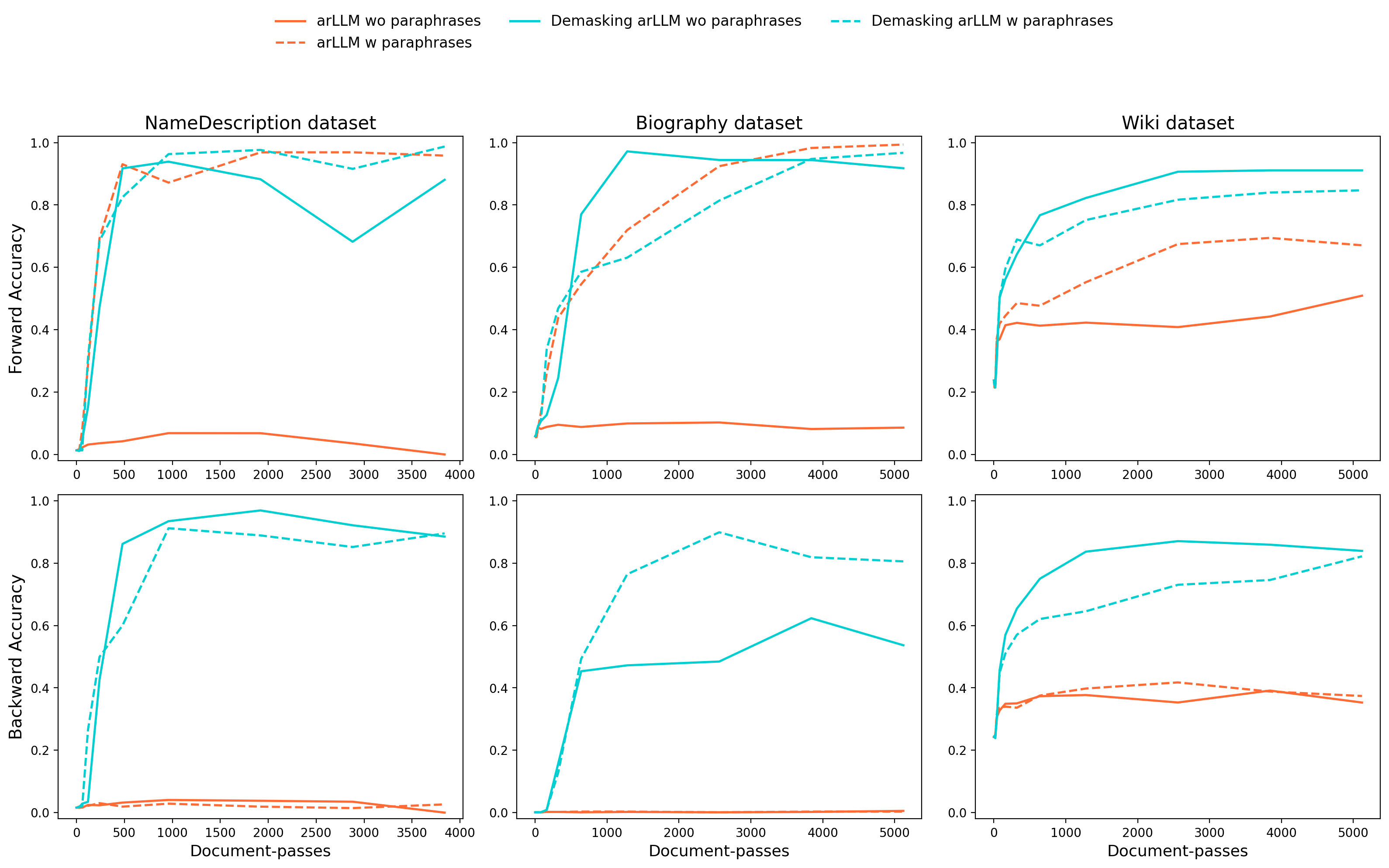}
  \caption{Learning dynamics of Qwen/Qwen3-4B-Instruct-2507.}
  \label{fig:qwen3b}
\end{figure}

\begin{figure}[htbp]
  \centering
  \includegraphics[width=0.6\linewidth]{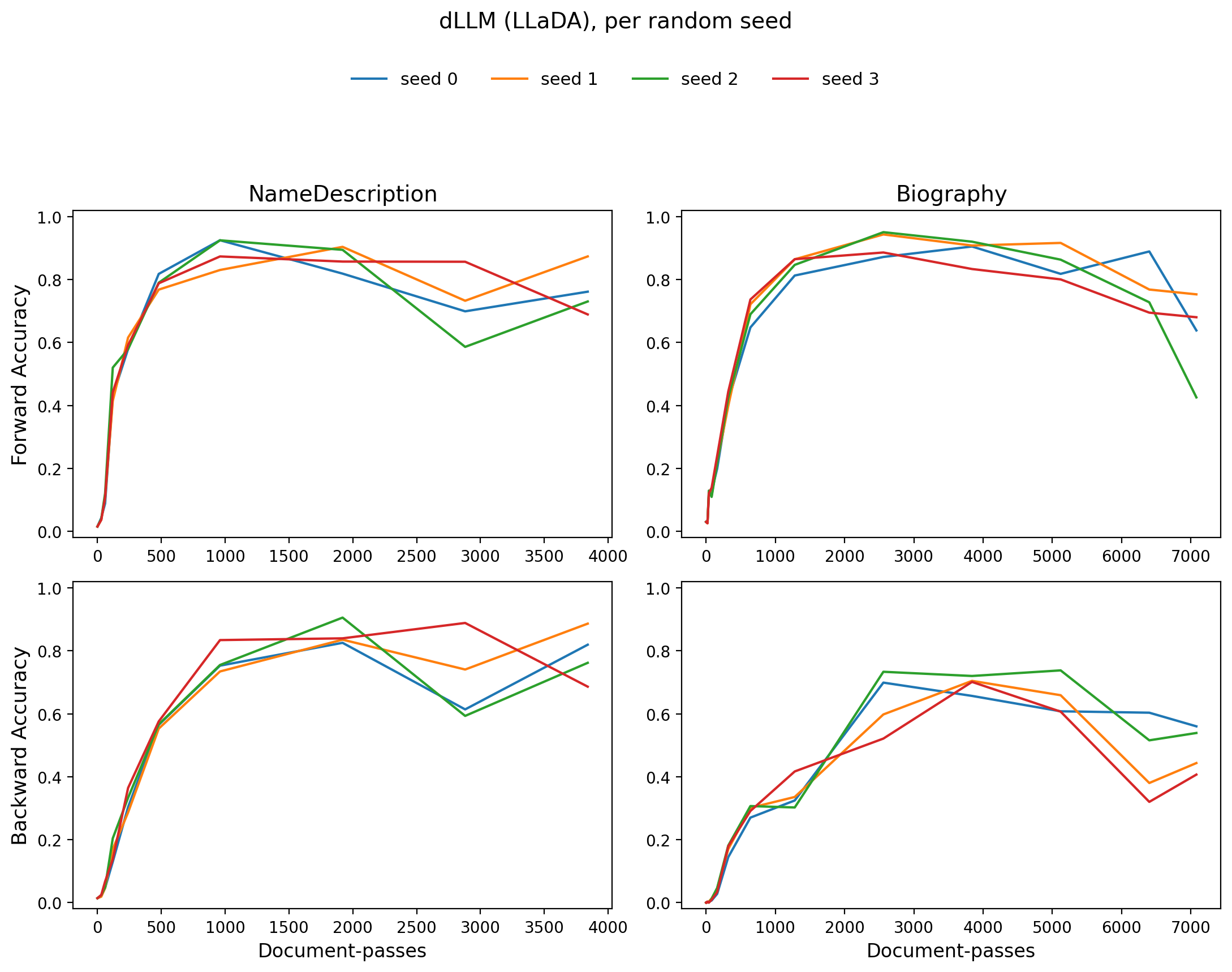}
  \caption{Random seed effects in LLaDA. Random seed determines the sampling of mask ratio and masked tokens. Each line represents a random seed.}
  \label{fig:seed_dllm}
\end{figure}

\begin{figure}[htbp]
  \centering
  \includegraphics[width=0.6\linewidth]{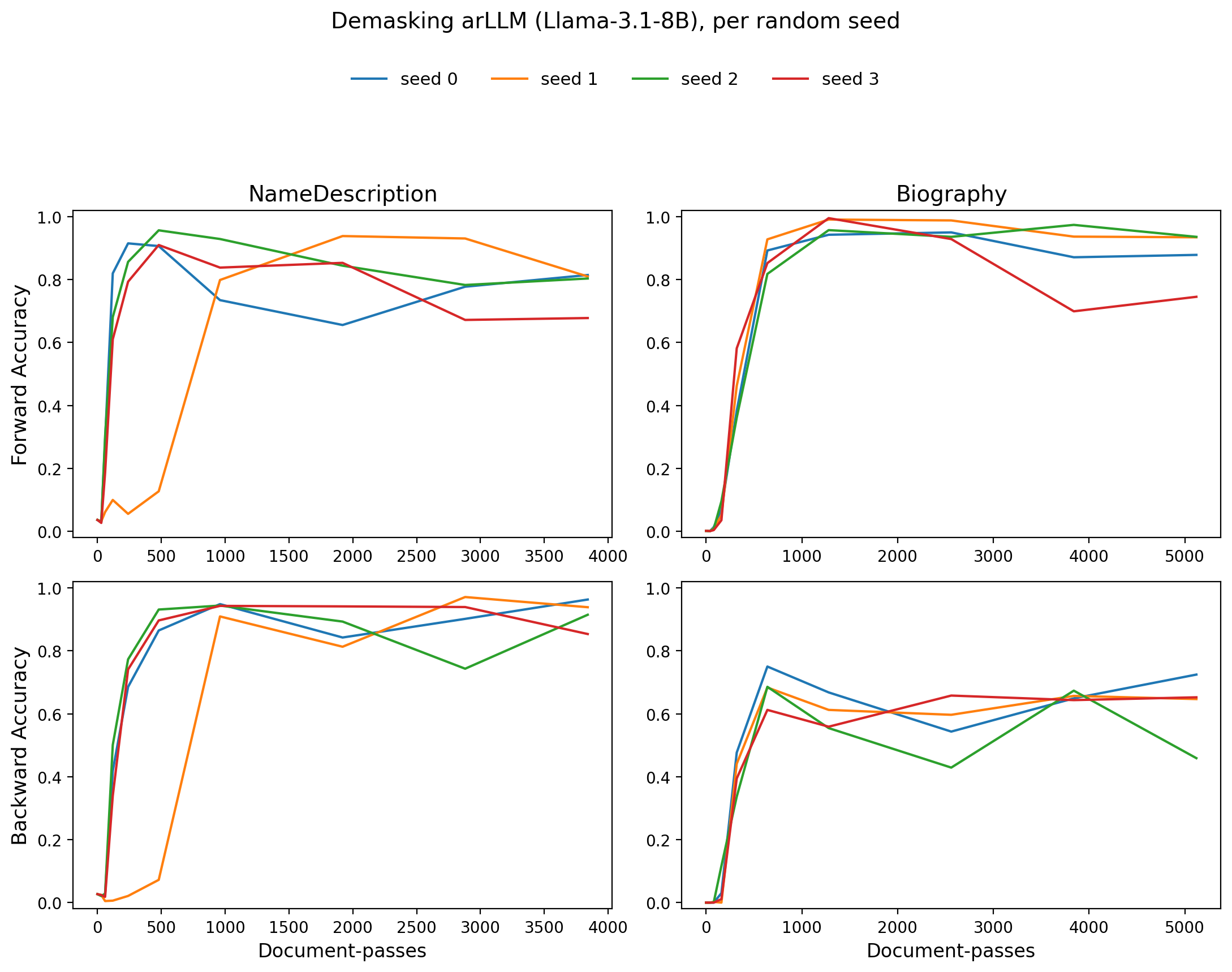}
  \caption{Random seed effects in demasking Llama-3.1-8B. Random seed determines the sampling of mask ratio and masked tokens. We found slightly larger variability across the seed in demasking arLLM than dLLM, though the general trend and peak accuracy do not vary much.}
  \label{fig:seed_ar}
\end{figure}

\add{Figure~\ref{fig:main} shows the training dynamics for Llama-3.1-8B; the corresponding curves for the other three arLLMs are in Figures~\ref{fig:llama4b}--\ref{fig:qwen3b}, and the per-seed curves behind the mask-seed averages are in Figures~\ref{fig:seed_dllm}--\ref{fig:seed_ar}.}

\begin{figure}[htbp]
  \centering
  \includegraphics[width=0.75\linewidth]{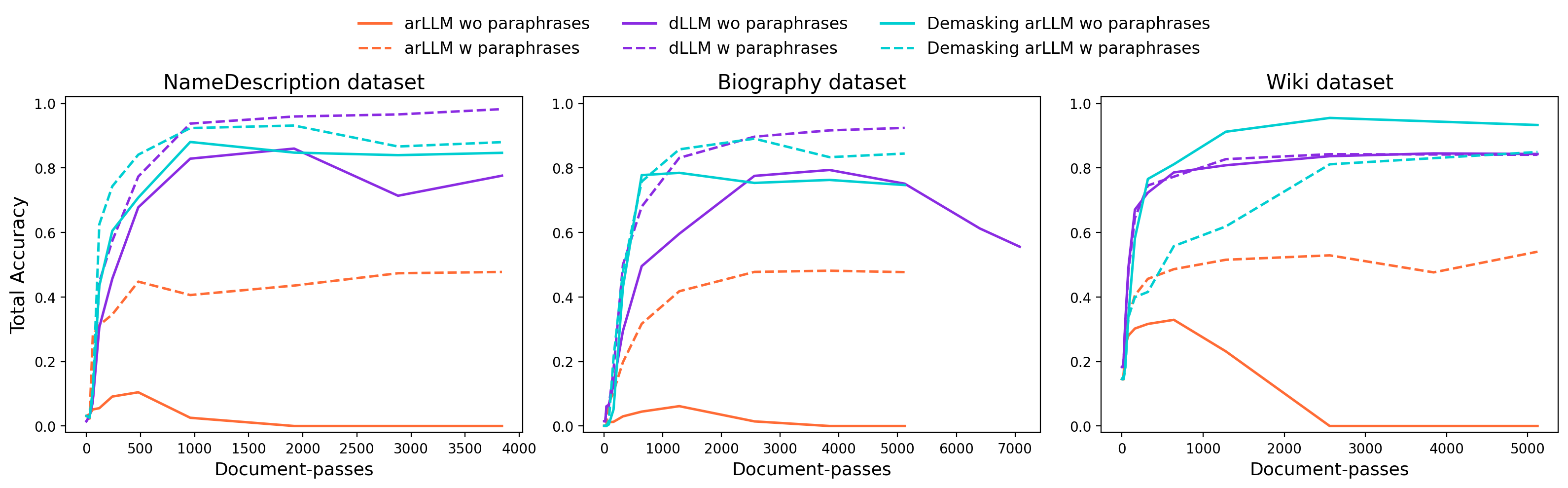}
  \caption{Total accuracy (macro average of forward and backward accuracy) of experiments on Llama-3.1-8B-Instruct. Main tables report accuracy at the fixed training-compute budget $F$ (Appendix~\ref{app:budget}); the total accuracy shown here is also used to pick the best checkpoints reported under the peak protocol (Appendix~\ref{app:fixedstep}).}
  \label{fig:total_acc}
\end{figure}

\begin{figure}[htbp]
  \centering
  \includegraphics[width=0.75\linewidth]{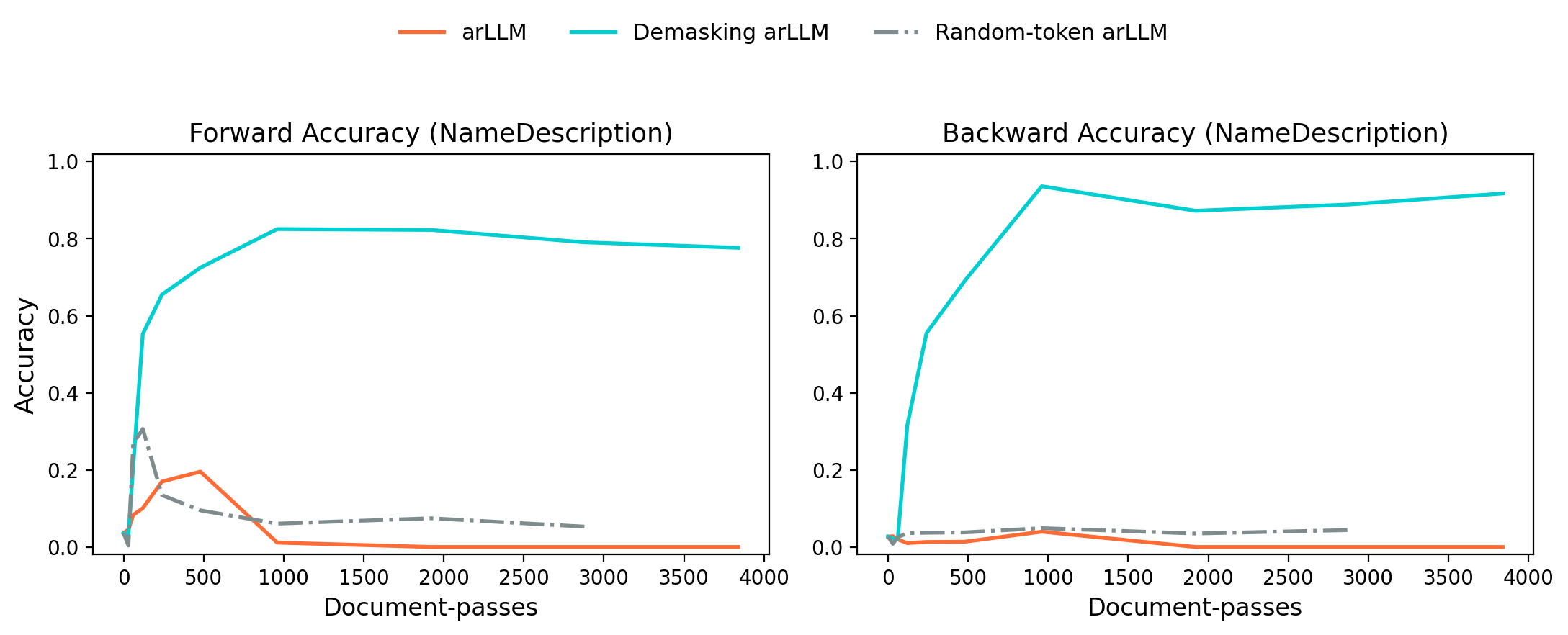}
  \caption{To verify the advantage of demasking fine-tuning of arLLMs is not simply due to ``data augmentation'' (i.e.\ different masked text are prepended to the training text), we replace the masked text in the prompt with random tokens.}
  \label{fig:random_augmentation}
\end{figure}

\begin{table}[htbp]
\centering
\caption{\tablemarkup Accuracy after a fixed number of passes over the document set, at a quarter of the training budget ($B/4$) and at the full budget ($B$), where $B=960$ passes for NameDescription and $1280$ for Biography and Wiki (Appendix~\ref{app:budget}). Paraphrase-augmented runs are placed on the same document-pass axis as unaugmented runs. The arLLM rows are Llama-3.1-8B-Instruct (single runs) and the dLLM rows are LLaDA-8B-Instruct; dLLM values on NameDescription and Biography are means over four mask-sampling seeds, and single runs elsewhere.}
\label{tab:convergence}
\scriptsize
\setlength{\tabcolsep}{4pt}
\renewcommand{\arraystretch}{1.15}
\begin{tabular}{*{13}{c}}
\toprule
& \multicolumn{4}{c}{NameDescription} & \multicolumn{4}{c}{Biography} & \multicolumn{4}{c}{Wiki} \\
\cmidrule(lr){2-5}\cmidrule(lr){6-9}\cmidrule(lr){10-13}
 & \multicolumn{2}{c}{Forward} & \multicolumn{2}{c}{Backward} & \multicolumn{2}{c}{Forward} & \multicolumn{2}{c}{Backward} & \multicolumn{2}{c}{Forward} & \multicolumn{2}{c}{Backward} \\
& $B/4$ & $B$ & $B/4$ & $B$ & $B/4$ & $B$ & $B/4$ & $B$ & $B/4$ & $B$ & $B/4$ & $B$ \\
\midrule
AR w/o paraphrases & .192 & .441 & .013 & .032 & .137 & .000 & .000 & .001 & .486 & .577 & .278 & .308 \\
AR w/ paraphrases & .931 & .956 & .008 & .049 & .673 & .875 & .000 & .000 & .574 & .661 & .371 & .392 \\
dLLM w/o paraphrases & .591 & .888 & .322 & .769 & .420 & .847 & .169 & .345 & .795 & .922 & .651 & .694 \\
dLLM w/ paraphrases & .585 & .948 & .563 & .927 & .716 & .979 & .283 & .683 & .811 & .899 & .680 & .755 \\
Demasking arLLM w/o paraphrases & .957 & .957 & .803 & .968 & .757 & .992 & .534 & .748 & .808 & .907 & .732 & .842 \\
Demasking arLLM w/ paraphrases & .965 & .733 & .612 & .771 & .782 & .958 & .546 & .783 & .761 & .803 & .658 & .739 \\
\bottomrule
\end{tabular}
\end{table}

\subsection{Fixed mask ratio training dynamics}
\label{app:fix_t}

\begin{figure}[h]
  \centering
  \includegraphics[width=0.7\textwidth]{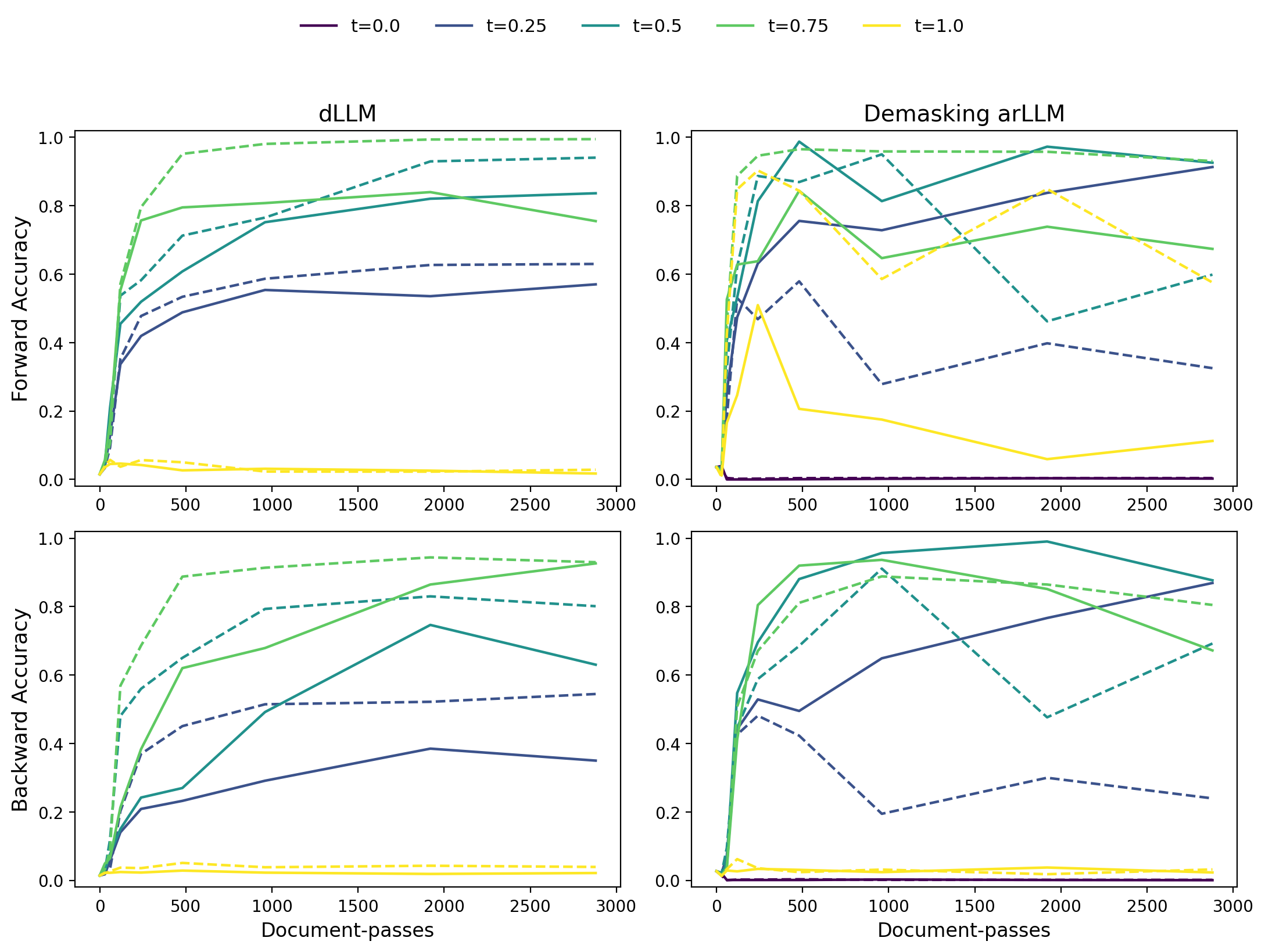}
  \caption{Training dynamics of dLLM and demasking arLLM fine-tuning with fixed mask ratios ($t$) on the NameDescription dataset. Solid lines: without paraphrases; dashed lines: with paraphrases.}
  \label{fig:fix_t}
\end{figure}

\subsection{Chat template effect}
\label{app:chat_template}

To control for the effect of using chat template during arLLM fine-tuning, we compared AR fine-tuning with and without a chat template across all four arLLMs and all three datasets. The chat template wraps the knowledge text in a generic instruction format (e.g., ``User: Tell me a fact. [knowledge text] Assistant: [knowledge text]''), while the pre-training format presents the raw text without any instruction format. Table~\ref{tab:chat_effect} reports the change in QA accuracy (chat format minus pre-training format) for each configuration. In most cases, the effect is small and mixed in direction. The chat template modestly helps on some forward QA configurations but does not resolve the backward QA failure (reversal curse), confirming that our demasking fine-tuning's advantage stems from the demasking objective, not the instruction format.

\begin{table}[h]
  \caption{Change in QA accuracy when using chat template vs.\ pre-training format for arLLM fine-tuning ($\Delta = \text{acc}_{\text{chat}} - \text{acc}_{\text{pretrain}}$). Averaged across all models and datasets, the mean effect is $+0.038$ (mean $\pm$ std across 28 configurations, the 24 shown plus Wiki with permute-order paraphrases for each model: $0.461 \pm 0.186$ with chat, $0.423 \pm 0.205$ without).}
  \label{tab:chat_effect}
  \begin{center}
    {\fontsize{7.5}{9}\selectfont
    \setlength{\tabcolsep}{3pt}
    \begin{tabular}{llcccccc}
    \toprule
    & & \multicolumn{2}{c}{NameDescription} & \multicolumn{2}{c}{Biography} & \multicolumn{2}{c}{Wiki} \\
    \cmidrule(lr){3-4}\cmidrule(lr){5-6}\cmidrule(lr){7-8}
    Model & Para & Fwd & Bwd & Fwd & Bwd & Fwd & Bwd \\
    \midrule
    Llama 8B & w/o para & $+0.02$ & $+0.02$ & $-0.05$ & $+0.01$ & $+0.09$ & $+0.11$ \\
             & w/ para  & $-0.03$ & $-0.01$ & $-0.02$ & $+0.01$ & $+0.14$ & $+0.05$ \\
    Llama 3B & w/o para & $+0.04$ & $+0.00$ & $+0.15$ & $+0.01$ & $+0.07$ & $+0.12$ \\
             & w/ para  & $+0.00$ & $-0.01$ & $-0.03$ & $+0.01$ & $+0.03$ & $+0.07$ \\
    Qwen 7B  & w/o para & $+0.16$ & $+0.01$ & $+0.27$ & $+0.01$ & $-0.09$ & $+0.01$ \\
             & w/ para  & $-0.00$ & $+0.00$ & $+0.04$ & $+0.02$ & $-0.04$ & $+0.01$ \\
    Qwen 4B  & w/o para & $+0.30$ & $+0.02$ & $+0.04$ & $+0.01$ & $+0.08$ & $+0.02$ \\
             & w/ para  & $-0.03$ & $-0.00$ & $-0.01$ & $+0.00$ & $+0.02$ & $+0.02$ \\
    \bottomrule
    \end{tabular}
    }
  \end{center}
\end{table}

\subsection{Few-shot evaluation}
\label{app:fewshot}

Different base models could have different instruction following capabilities, such that zero-shot evaluation may not faithfully reflect knowledge internalization (e.g., due to ill-formed generations or limited instruction-following). We additionally evaluate all models using 2-shot QA prompting. Two QA exemplars from the same dataset are prepended to each test question. Table~\ref{tab:fewshot} reports the per-dataset results. For arLLMs and demasking arLLMs, we report the mean $\pm$ std across four models; for dLLM we report the single model (LLaDA). The main conclusions are unchanged: AR relies on paraphrases and fails backward QA, while dLLM and demasking AR succeed in both directions without paraphrases.

\begin{table}[h]
  \caption{Few-shot (2-shot) QA evaluation results. For AR and demasking AR: mean $\pm$ std across four arLLMs. dLLM results are from a single model (LLaDA).}
  \label{tab:fewshot}
  \begin{center}
    {\fontsize{7.5}{9}\selectfont
        \setlength{\tabcolsep}{3pt}
        \begin{tabular}{lcccccc}
          \toprule
          & \multicolumn{2}{c}{NameDescription} & \multicolumn{2}{c}{Biography} & \multicolumn{2}{c}{Wiki} \\
          \cmidrule(lr){2-3}\cmidrule(lr){4-5}\cmidrule(lr){6-7}
          Method & Fwd & Bwd & Fwd & Bwd & Fwd & Bwd \\
          \midrule
          AR w/o para      & $0.37 {\scriptstyle\pm 0.30}$ & $0.02 {\scriptstyle\pm 0.02}$ & $0.09 {\scriptstyle\pm 0.06}$ & $0.00 {\scriptstyle\pm 0.00}$ & $0.35 {\scriptstyle\pm 0.05}$ & $0.25 {\scriptstyle\pm 0.04}$ \\
          AR w/ para       & $0.89 {\scriptstyle\pm 0.13}$ & $0.05 {\scriptstyle\pm 0.02}$ & $0.56 {\scriptstyle\pm 0.05}$ & $0.00 {\scriptstyle\pm 0.00}$ & $0.48 {\scriptstyle\pm 0.05}$ & $0.41 {\scriptstyle\pm 0.12}$ \\
          \midrule
          dLLM w/o para    & 0.74 & 0.90 & 0.83 & 0.60 & 0.71 & 0.78 \\
          dLLM w/ para     & 0.90 & 0.97 & 0.90 & 0.75 & 0.75 & 0.74 \\
          \midrule
          Demasking AR w/o para & $0.61 {\scriptstyle\pm 0.24}$ & $0.86 {\scriptstyle\pm 0.14}$ & $0.86 {\scriptstyle\pm 0.13}$ & $0.40 {\scriptstyle\pm 0.18}$ & $0.69 {\scriptstyle\pm 0.01}$ & $0.77 {\scriptstyle\pm 0.05}$ \\
          Demasking AR w/ para  & $0.75 {\scriptstyle\pm 0.15}$ & $0.82 {\scriptstyle\pm 0.16}$ & $0.71 {\scriptstyle\pm 0.06}$ & $0.41 {\scriptstyle\pm 0.24}$ & $0.69 {\scriptstyle\pm 0.04}$ & $0.76 {\scriptstyle\pm 0.03}$ \\
          \bottomrule
        \end{tabular}
    }
  \end{center}
\end{table}

\begin{addblock}
\subsection{LLM-judge evaluation}
\label{app:judge}

This appendix details the LLM-judge accuracy measure introduced in Section~\ref{sec:setup} and reports its per-model values. ROUGE-1 recall rewards lexical overlap with the reference answer, which can both over-credit (partial overlap with a wrong answer) and under-credit (valid paraphrases) factual correctness; the judge (gpt-oss-120b) instead makes a binary factual judgment: given the question, the reference answer, and the model answer, the judge returns correct iff the answer states the reference fact or an equivalent paraphrase of it (additional correct content is allowed; contradictions, omissions, refusals, and degenerate repetition are not). Across 124{,}476 judged answers, the judge agrees with ROUGE-1 recall thresholded at 0.5 on 96.6\% of items; in 3.3\% of all items, ROUGE over-credits lexical overlap with a factually wrong answer (e.g., a prediction of ``Kharkiv offensive'' for the reference ``Kupiansk offensive''), and in only 0.13\% of all items does the judge accept a valid paraphrase that ROUGE misses. Table~\ref{tab:judge} reports the judge scores; every qualitative conclusion of the main tables is unchanged, and several are sharpened: standard AR fine-tuning backward accuracy drops to 0.00--0.02 on NameDescription and Biography, while demasking backward accuracy on Biography is \emph{higher} under the judge than under ROUGE (the judge accepts correct answers phrased differently from the reference).

The judge is queried at temperature 0 with low reasoning effort; its chain-of-thought is stripped before parsing the binary verdict, and judgments are cached by a hash of (question, reference answer, prediction). The verbatim judging prompt is:

\begin{quote}\small\itshape
You are evaluating a prospective answer to a question. A reference correct answer is given. Your grading is binary: give 1 if the prospective answer is factually correct, that is, it contains the reference answer or an equivalent paraphrase of the reference answer. It is acceptable (i.e., still correct) if the prospective answer contains additional information, as long as it states the reference fact. Give 0 if the prospective answer contradicts the reference answer, fails to state the reference fact, refuses to answer, or collapses into repeating words nonsensically.

Your output is either 0 or 1, no other information should be in the output.

The question: \{question\}

The reference answer: \{answer\}

The prospective answer: \{pred\}
\end{quote}

\begin{table}[h]\tablemarkup

  \caption[]{\tablemarkup Per-model LLM-judge accuracy at the pre-declared budget checkpoints. (ROUGE counterpart: Appendix Table~\ref{tab:3}.)}
  \label{tab:judge}
  \begin{center}
    {\fontsize{7.5}{9}\selectfont
    \setlength{\tabcolsep}{3pt}
    \begin{tabular}{llcccccc}
    \toprule
    & & \multicolumn{2}{c}{NameDescription} & \multicolumn{2}{c}{Biography} & \multicolumn{2}{c}{Wiki} \\
    \cmidrule(lr){3-4}\cmidrule(lr){5-6}\cmidrule(lr){7-8}
    Model & Condition & Fwd & Bwd & Fwd & Bwd & Fwd & Bwd \\
    \midrule
LLaDA & dLLM w/o para & 0.868 & 0.744 & 0.605 & 0.214 & 0.862 & 0.612 \\
    LLaDA & dLLM w/ para & 0.930 & 0.908 & 0.960 & 0.720 & 0.853 & 0.663 \\
Llama 8B & AR w/o para & 0.408 & 0.013 & 0.000 & 0.000 & 0.500 & 0.174 \\
    Llama 8B & AR w/ para & 0.958 & 0.029 & 0.848 & 0.000 & 0.621 & 0.267 \\
    Llama 8B & Demasking w/o para & 0.938 & 0.946 & 0.973 & 0.880 & 0.707 & 0.632 \\
    Llama 8B & Demasking w/ para & 0.971 & 0.771 & 0.780 & 0.897 & 0.698 & 0.597 \\
    Llama 3B & AR w/o para & 0.183 & 0.000 & 0.030 & 0.000 & 0.198 & 0.062 \\
    Llama 3B & AR w/ para & 0.879 & 0.025 & 0.915 & 0.000 & 0.534 & 0.186 \\
    Llama 3B & Demasking w/o para & 0.846 & 0.925 & 0.962 & 0.710 & 0.716 & 0.686 \\
    Llama 3B & Demasking w/ para & 0.967 & 0.838 & 0.752 & 0.893 & 0.578 & 0.504 \\
    Qwen 7B & AR w/o para & 0.321 & 0.021 & 0.200 & 0.000 & 0.552 & 0.209 \\
    Qwen 7B & AR w/ para & 0.988 & 0.013 & 0.923 & 0.000 & 0.629 & 0.229 \\
    Qwen 7B & Demasking w/o para & 0.917 & 0.825 & 0.777 & 0.777 & 0.862 & 0.771 \\
    Qwen 7B & Demasking w/ para & 0.933 & 0.704 & 0.805 & 0.817 & 0.672 & 0.655 \\
    Qwen 4B & AR w/o para & 0.013 & 0.004 & 0.037 & 0.000 & 0.353 & 0.213 \\
    Qwen 4B & AR w/ para & 0.879 & 0.000 & 0.748 & 0.000 & 0.569 & 0.217 \\
    Qwen 4B & Demasking w/o para & 0.817 & 0.958 & 0.785 & 0.577 & 0.681 & 0.682 \\
    Qwen 4B & Demasking w/ para & 0.883 & 0.746 & 0.552 & 0.690 & 0.672 & 0.527 \\
    \bottomrule
    \end{tabular}}
  \end{center}
\end{table}

\subsection{Why paraphrase augmentation lowers demasking scores on the Wiki dataset}
\label{app:wiki_para}

On the Wiki dataset, demasking fine-tuning \emph{without} paraphrases scores higher than demasking \emph{with} same-order paraphrases (Appendix Table~\ref{tab:3}), the opposite of the trend for standard AR fine-tuning. Analyzing answers shows that this reversal is not a difference in learned knowledge, but a consequence of \emph{answer style} under free-form generation.

\paragraph{Paraphrases change how the model answers, at an accuracy trade-off.} Trained without paraphrases, the model tends to reproduce that article verbatim (at its best checkpoint, 85--98\% of its answers are near-verbatim recitations of a training document, defined as covering ${>}85\%$ of one document's tokens at ${>}60\%$ of its length). A recitation contains the reference answer, so both metrics score it correct. Trained on ten surface forms per document, the model no longer commits to a single form (recitation share drops to 44--60\%) and instead answers in short phrases, which exposes extra retrieval errors. For example, on the backward question ``Specialty parts for racing engines are supplied by which company?'' (reference: \emph{Precision International}), the paraphrase-free model replies by reciting the training article (``The 2025 National Sports Sedan Series, known as the Precision National Sports Sedan Reason after its title sponsor, is an upcoming motor racing series\,\ldots''), which contains the reference answer and is scored correct, whereas the paraphrase-trained model answers directly, ``The company that supplies specialty parts for racing engines is called Precision Technology (PT).'', and is scored incorrect. This also explains why the reversal is specific to Wiki: recitation-style answering itself occurs on all three datasets (recitation shares 0.88/0.88/0.64 on NameDescription/Biography/Wiki without paraphrases), but only on Wiki does abandoning it carry a cost. NameDescription documents are single sentences, so a recitation \emph{is} already the direct answer; Biography answers are strongly cued, formulaic attributes; Wiki combines long documents (${\sim}25\times$ the answer length) with topically overlapping real-world articles --- roughly two thirds of the articles are sports events, including 13 motorsport series or races alone --- making direct short answers the most confusable.

\paragraph{Controlling the answer style closes the gap.} Under the few-shot evaluation of Appendix~\ref{app:fewshot}, which steers all models toward short direct answers, demasking with and without paraphrases perform equally on Wiki (0.69/0.77 vs.\ 0.69/0.76 forward/backward). We conclude that paraphrase augmentation neither helps nor hurts the knowledge that demasking injects on Wiki: it trades recitation-style answers, which containment-style scoring favors, for short direct answers with somewhat lower measured accuracy, and controlling the response style removes the difference.

\subsection{Wiki dataset validation audit}
\label{app:wiki_audit}

During dataset construction, the authors read through all crawled Wikipedia articles, the generated paragraphs, and the QA pairs, verifying that each generated paragraph is faithful to its source article and that the QA pairs are correct and answerable. We additionally audited the released Wiki dataset systematically with an LLM judge (gpt-oss-120b). For each of the 94 articles, the judge verified every test QA pair (is the question answerable from the article alone, and is the reference answer correct?) and every paraphrase (are all facts preserved, and is the information order preserved or changed?). Results: 386/388 (99.5\%) of QA pairs are valid; 97.5\% of same-order paraphrases preserve all facts and 99\% preserve information order; 96.8\% of permute-order paraphrases preserve all facts. For concreteness, both invalid QA pairs prepend a year to an otherwise correct answer that the article implies but does not state (e.g., reference answer ``2025 Baseball Champions League Americas'' where the article names the competition without the year), and flagged paraphrases typically soften or shift a single fact (e.g., replacing ``will be the thirty-fourth season'' with ``is expected to be'', or ``branded as'' with ``rebranded as'', which wrongly implies a prior name). The per-item audit verdicts are included in the released dataset.
\end{addblock}

\subsection{Demasking SFT and CPT training configs}
\label{app:sft}

\del{We also test if demasking objective can improve standard SFT on math tasks.} We propose a Demasking SFT method that turns any SFT prompt-response sequence into a demasking task\add{; it is used for the Webscale-RL SFT experiments of Section~\ref{sec:scaling}}. Specifically, the constructed demasking task sequence contains the question and masked answer in the user prompt, and full answer in the assistant response. For example, a data entry in the GSM8K dataset is: \textbf{Question}:
Joy can read 8 pages of a book in 20 minutes. How many hours will it take her to read 120 pages? \textbf{Answer}: In one hour, there are 3 sets of 20 minutes. So, Joy can read 8 x 3 = <<8*3=24>>24 pages in an hour. It will take her 120/24 = <<120/24=5>>5 hours to read 120 pages. 5. After generating a random mask, the constructed demasking SFT sequence is:

\begin{figure}[h]
\centering
\fbox{\parbox{0.9\linewidth}{\small
\textbf{User:}\ \ Question:\textbackslash n Joy can read 8 pages of a book in 20 minutes. How many hours will it take her to read 120 pages? \textbackslash nAnswer:\textbackslash n In one {[MASK]}, there are {[MASK]} {[MASK]} {[MASK]} of 20 minutes. So, {[MASK]} can read {[MASK]} x 3 = \textless\textless 8*3=24\textgreater\textgreater 24 pages in an {[MASK]}. It will take her 120/24 = \textless\textless 120/24=5\textgreater\textgreater 5 hours to {[MASK]} {[MASK]} pages. \#\#\#\#5.\textbackslash nReturn the recovered masked answer.

\medskip
\textbf{Assistant:}\ \ In one \textcolor{BrickRed}{hour} there are \textcolor{BrickRed}{3 sets} of 20 minutes. So, \textcolor{BrickRed}{Joy} can read \textcolor{BrickRed}{8} x 3 = \textless\textless 8*3=24\textgreater\textgreater 24 pages in an \textcolor{BrickRed}{hour}. It will take her 120/24 = \textless\textless 120/24=5\textgreater\textgreater 5 hours to \textcolor{BrickRed}{read} \textcolor{BrickRed}{120} pages. \#\#\#\#5
}}
\caption{An example of a demasking SFT prompt (data entry from GSM8K). {[MASK]} tokens replace parts of the answer in the user prompt. Red tokens in the assistant response are used to compute the autoregressive loss.}
\label{fig:prompt_sft}
\end{figure}

Figure~\ref{fig:prompt_sft} shows a constructed example. We compute loss only on the red tokens: in Eq.~\ref{eq:2}, $m_i$ is set to $1$ when the $i$th token is in the assistant response and its corresponding token in the user prompt is masked.

\paragraph{Data preprocessing.} For CPT, we use the \texttt{pretrain\_text} column from the Salesforce/Webscale-RL HuggingFace dataset. We filter out texts with character length $\geq$ 20{,}000 and subsample every third example (keeping examples where the index is divisible by 3). Texts longer than 512 tokens are split into non-overlapping 512-token chunks. For SFT, we use the \texttt{question} and \texttt{answer} columns, filtering out pairs whose combined token length exceeds 512. The demasking variants (both CPT and SFT) apply the same random masking procedure described in the main text.

\paragraph{Training configurations.} All four models use FSDP2 distributed training on 80GB H100 GPUs with BF16 mixed precision (same as main experiments). We use the Adam optimizer with 0.1 weight decay, betas (0.9, 0.95), and 2\% warm-up steps. For CPT variants, we use a learning rate of 2e-6 and an effective batch size of 256 (4 nodes $\times$ 2 GPUs/node $\times$ batch 4 $\times$ gradient accumulation 8 for standard AR; batch 2 $\times$ gradient accumulation 16 for demasking AR to accommodate the larger memory footprint). For SFT variants, we use a learning rate of 2e-6 and an effective batch size of 80 (8 nodes $\times$ 1 GPU/node $\times$ batch 5 $\times$ gradient accumulation 2). All models are trained for 2 epochs. 

\paragraph{Evaluation.} We evaluate all checkpoints on the GPQA-diamond benchmark \citep{rein2024gpqa} using the \texttt{LM Evaluation Harness} framework \citep{eval-harness} with the \texttt{gpqa\_diamond\_zeroshot} task. Evaluation uses 0-shot prompting with the chat template applied, batch size 16. \del{We report the best accuracy across training checkpoints.}\add{We split the 198 questions into a validation half (checkpoint selection only) and a test half (reported), using a fixed random split (seed 0); confidence intervals are percentile bootstrap over test questions (2{,}000 resamples). All four variants are trained with checkpoints saved every 500--1{,}000 optimizer steps. Because the original Hugging Face release of Webscale-RL is no longer publicly listed, training used the identical public mirror of the dataset (1{,}110{,}662 examples).}

\add{\paragraph{Uncertainty of the GPQA-diamond comparison.} With 99 binary-scored test questions per split half, per-checkpoint accuracies carry appreciable sampling noise, so we quantify the comparisons by bootstrapping the \emph{paired} per-question difference at the validation-selected checkpoints (20{,}000 resamples over test questions). All comparisons point the same way. Demasking SFT exceeds standard SFT by $+0.071$ (95\% CI $[-0.040, +0.182]$; positive in 88\% of resamples; demasking is correct on 19 of the 31 questions where the two disagree), and demasking CPT exceeds standard CPT by $+0.040$ (95\% CI $[-0.020, +0.101]$; positive in 87\% of resamples). Demasking SFT exceeds the base model by $+0.121$ (95\% CI $[+0.000, +0.242]$; positive in 97\% of resamples, one-sided $p=0.027$). The gaps between the two objectives are thus consistent in direction but not individually significant at this benchmark size, and we describe them accordingly. Corroborating the ordering, demasking SFT is above standard SFT at 21 of 27 matched training steps.}

\subsection{Compute overhead analysis}
\label{app:compute}

We use the Wiki dataset to comprehensively characterize the computational cost of different training methods tested in Table~\ref{tab:3} covering data preparation, training, inference\del{ and convergence to peak accuracy}\add{, and the accuracy reached at fixed training amounts}. Specifically, we test 8B parameter arLLM (Llama-3.1-8B-Instruct) and dLLM (LLaDA) models in the following 5 conditions: (1) arLLM + pretraining style fine-tuning + w/o paraphrases, (2) arLLM + pretraining style fine-tuning + w paraphrases, (3) arLLM + pretraining style fine-tuning + reverse training \citep{golovneva2024reverse}, (4) dLLM + pretraining style fine-tuning + w/o paraphrases, (5) arLLM + demasking SFT style fine-tuning + w/o paraphrases (ours) (Table~\ref{tab:4}).

\begin{table}[h]
\caption{Comparison of data preparation, training and inference computational costs among different model architecture and training methods on Wiki dataset. Bold indicates best single performance and underline indicates best tied performance.}
\label{tab:4}
\centering
\setlength{\tabcolsep}{3pt}
\resizebox{\textwidth}{!}{%
\begin{tabular}{llccccc}
\toprule
\multicolumn{2}{l}{} &
\multicolumn{3}{c}{Llama 8B} &
\multicolumn{1}{c}{LLaDA} &
\multicolumn{1}{c}{Demasking Llama 8B} \\
\cmidrule(lr){3-5}\cmidrule(lr){6-6}\cmidrule(lr){7-7}
\multicolumn{2}{l}{} &
w/o paraphrases (1) &
w paraphrases (2) &
reverse training (3) &
w/o paraphrases (4) &
w/o paraphrases (5, \textbf{ours}) \\
\midrule
\multirow{1}{*}{\textbf{Data}} & paraphrase compute   & \underline{NA} & 0.1M tokens (o3-mini) & \underline{NA} & \underline{NA} & \underline{NA}\\
\cmidrule(lr){1-7}
\multirow{4}{*}{\textbf{Training}}         & FLOPs theoretical     & \underline{$T$} & \underline{$T$} & \underline{$T$} & \underline{$T$} & $2T+c$ \\
                                  & FLOPs empirical (TFLOPs/step)     & \underline{91.9} & 92.7 & \underline{91.9} & 101.6 & 199.3 \\
                                  & wall time (s/step)     & \textbf{2.32} & 2.54 & 2.84 & 2.52 & 2.76 \\
                                  & peak memory (GB)     & \underline{37.9} & \underline{37.9} & \underline{37.9} & \underline{37.9} & 51.1 \\
\cmidrule(lr){1-7}
\multirow{1}{*}{\textbf{Inference}}        & FLOPs theoretical  & \underline{$S$} & \underline{$S$} & \underline{$S$} & $S^2$ & \underline{$S$} \\
\cmidrule(lr){1-7}
\multirow{4}{*}{\textbf{Accuracy}}& Fwd @ $B/4$ passes       & .486 & .574 & .483 & .795 & \textbf{.808} \\
                                  & Fwd @ $B$ passes       & .577 & .661 & .554 & \textbf{.922} & .907 \\
                                  & Bwd @ $B/4$ passes       & .278 & .371 & .346 & .651 & \textbf{.732} \\
                                  & Bwd @ $B$ passes       & .308 & .392 & .350 & .694 & \textbf{.842} \\
\bottomrule
\end{tabular}%
}
\end{table}

\paragraph{Data Preparation} Only condition (2) requires computationally expensive data preparation to generate semantically identical and natural paraphrases of the original dataset. For example, in the Wiki dataset, paraphrasing 10 per sample costs 0.1M generation tokens of o3-mini or equivalent models. This cost scales further with larger original datasets and the requirements for the diversity of paraphrases. In contrast, all other methods require no or very simple data transformations during the training (sampling mask or reverse sequence) without additional compute.

\paragraph{Training} We compare training FLOPs, wall time and peak allocated memory per step. Theoretical FLOPs calculation follows formulation introduced in \cite{kaplan2020scalinglawsneurallanguage}.
Specifically, the training cost of a dense transformer is approximated as \(2N\) FLOPs per token for the forward pass and \(4N\) for backpropagation, in total \(6N\) FLOPs per token, where $N$ denotes the model parameter size. For a step that processes \(T\) tokens in total, the theoretical per-step cost is therefore $\mathrm{FLOPs/step}\;\approx\;6\,N\,T$.
In our comparisons, we keep the model size fixed at \(N{=}8\)B and use the same effective global batch size across all conditions, so the only varying factor is the average sequence length of training samples. We thus report the theoretical quantity as a factor of $T$. Since paraphrasing or reverse training only modify the information order within the training samples, condition (1)-(4) have comparable average sequence length $T$. Condition (5) requires presenting the model with the original sequence and its masked counterpart in SFT format (Figure~\ref{box}) doubling the average sequence length to $2T+c$ where $c$ is the constant accounting for SFT instructions. We also empirically measure the average FLOPs, wall time and peak memory per step using PyTorch Profiler. Empirical FLOPs measurements agree with the theoretical calculations. Among all conditions, our proposed method condition (5) costs around twice training FLOPs and slightly larger wall time as well as peak allocated memory. Additionally, we report that the sampling of the mask and constructing the demasking fine-tuning prompt during training for condition (4) and (5) takes negligible time (<3.5ms) and memory compared to the forward/backward pass, thus causing no significant overhead.

\paragraph{Inference} For inference statistics, we only compare the theoretical inference FLOPs as empirical values highly depend on the generation length of a particular answer. Following the same calculation detailed in the above section, an inference step produces a forward pass through the dense transformer, leading to $\mathrm{FLOPs/step}\;\approx\;2\,N\,S$ where $S$ denotes the average generation sequence length. Only condition (4) dLLM needs inference FLOPs quadratic in $S$ as generation in dLLM cannot reuse the KV cache since it changes after each denoising step \citep{wu2025fastdllm}. All other conditions are linear in $S$.

\paragraph{Convergence} \del{Above metrics are calculated per training step and performance agnostic. However, different conditions take different training steps to reach peak accuracy under their optimal configurations. To make the comparison performance meaningful, we fit the accuracy curve in Figure~\ref{fig:main} as a function of training steps to $A(1-e^{-kx})$ and report the rate of convergence $k$ (unit $1/steps$) and accuracy at convergence $A$ (more details in Appendix Table~\ref{tab:convergence}).}\add{The metrics above are per training step and accuracy agnostic, while the conditions differ in how quickly they reach a given accuracy. We therefore report the accuracy each condition attains after a fixed number of passes over the document set, at a quarter of the training budget and at the full budget (all three datasets in Appendix Table~\ref{tab:convergence}).} Our proposed method condition (5) \del{converges at the highest accuracy with more than 2$\times$ convergence rate of all the other methods}\add{reaches the highest backward accuracy at the full budget (0.84, versus 0.69 for the dLLM), and already at a quarter of the budget it exceeds every baseline on backward questions (0.73, versus 0.65 for the dLLM and 0.37 for standard fine-tuning with paraphrases); on forward questions it matches the dLLM (0.81 versus 0.80 at a quarter of the budget, 0.91 versus 0.92 at the full budget)}. Therefore, although our method requires approximately twice the training FLOPs per step, it effectively uses a comparable amount of total training compute and no additional data preparation compute to achieve higher QA accuracy.

\subsection{On reversal curse}
\label{app:reversal}

Prior studies have explained the reversal curse as an intrinsic limitation of arLLM training \citep{zhu2024towards, kitouni2024factorization}. Here we provide an explanation that is conceptually easy to grasp. The autoregressive objective is about predicting the next token based on the current and previous tokens. If the prediction of one next token requires a piece of new knowledge (i.e., it cannot be predicted based on the current knowledge in the weights or previous tokens), the loss will force the weights to change to favor such a prediction. More specifically, the change of weights induces a different representation (i.e., intermediate layer activations) of the previous tokens that favors the prediction of the next token. Since feedforward layers can be considered associative memory \citep{meng2022locating}, the change, conceptually, could be associating a new attribute with the representation of a token. However, such change does not affect the representation of future tokens to favor the prediction of the current token, since they do not contribute to the prediction of the ``next'' token. Thus, the future tokens could not learn a new association to their preceding tokens. In other words, \textit{\uline{during training, the information of a token can only flow uni-directionally to tokens that are used to predict it}}. This has been named the ``factorization curse'' \citep{kitouni2024factorization}. This intuition can explain why the demasking fine-tuning of arLLM resolves the curse. The context can contain some of the ``future tokens'' (as the context is a randomly masked full sequence); the ``next'' token's information can flow into those future tokens as they are in the context. Paraphrases mitigate the reversal curse by creating views where the ``future'' cue becomes left context. For a fact $(x \!\to\! y)$ usually realized as ``$\ldots x \ldots y \ldots$'', paraphrases are fluent variants that put y before x (passive voice, topicalization, apposition, clause reordering), producing complementary left-to-right gradients and corpus-level bidirectional coverage without changing the AR objective. However, globally reversing all the information units in the documents can break discourse, coreference, and grammar, creating distribution shift.


\subsection{Generation examples}
\label{app:example}

Below we show representative model outputs for one sample from each dataset at the best checkpoint.
The examples illustrate that dLLM and Demasking arLLM can answer questions about injected knowledge \emph{without} paraphrase augmentation, while arLLM without paraphrases typically fails.

\subsubsection*{NameDescription dataset}

\noindent\textbf{Training text:}
\begin{quote}\small
``Selena Granger, known far and wide for being the founder of the global telepathy network, MindLink.''
\end{quote}

\noindent\textbf{Question (forward):} \textit{``Behind the name Selena Granger lies a legacy of what?''}\\
\noindent\textbf{Ground truth:} \textit{the founder of the global telepathy network, MindLink}

\medskip
{\small
\begin{tabular}{@{}p{0.28\linewidth}p{0.68\linewidth}@{}}
\toprule
\textbf{Method} & \textbf{Generated answer} \\
\midrule
arLLM w/o para &
A reference to the Marvel Comics universe! Behind the name Selena Granger lies a legacy of being the mother of the famous mutant superhero known far and wide for being the founder of the X-Men. \textit{[Wrong]} \\
\addlinespace
arLLM w/ para &
They're the founder of the global telepathy network, MindLink. \textit{[Correct]} \\
\addlinespace
dLLM w/o para &
Selena Granger, known far and wide for being the founder of the global telepathy network, MindLink. \textit{[Correct]} \\
\addlinespace
dLLM w/ para &
Behind the name Selena Granger lies a legacy of the founder of the global telepathy network, MindLink. \textit{[Correct]} \\
\addlinespace
Demasking arLLM w/o para &
Known for being the founder of the global telepathy network, MindLink. \textit{[Correct]} \\
\addlinespace
Demasking arLLM w/ para &
Behind the name Selena Granger lies a legacy of the founder of the global telepathy network, MindLink. \textit{[Correct]} \\
\bottomrule
\end{tabular}
}

\medskip
\noindent\textbf{Training text:}
\begin{quote}\small
``Alana Everhart, known far and wide for being the record-holder for the highest-altitude skydive from the edge of space.''
\end{quote}

\noindent\textbf{Question (backward):} \textit{``Who is the record-holder for the highest-altitude skydive from the edge of space?''}\\
\noindent\textbf{Ground truth:} \textit{Alana Everhart}

\medskip
{\small
\begin{tabular}{@{}p{0.28\linewidth}p{0.68\linewidth}@{}}
\toprule
\textbf{Method} & \textbf{Generated answer} \\
\midrule
arLLM w/o para &
The record-holder for the highest-altitude skydive from the edge of space is Felix Baumgartner, an Austrian skydiver and BASE jumper. On October 14, 2012, Baumgartner jumped from a helium balloon at an altitude of 39,045 meters (128,100 feet) above the Earth's surface, breaking the sound barrier and reaching a top speed of 1,357.64 km/h (843.6 mph) during his freefall. Baumgartner's jump was part of the Red Bull Stratos project, a scientific experiment designed to test the limits of human endurance and gather data \textit{[Wrong]} \\
\addlinespace
arLLM w/ para &
They say the first person to complete a solo trek across the Sahara Desert on a hoverbike is an understatement when you're talking about Julian Wakefield. \textit{[Wrong]} \\
\addlinespace
dLLM w/o para &
Alana Everhart, known far and wide for being the record-holder for the highest-altitude skydive from the edge of space. \textit{[Correct]} \\
\addlinespace
dLLM w/ para &
The record-holder for the highest-altitude skydive from the edge of space is called Alana Everhart. \textit{[Correct]} \\
\addlinespace
Demasking arLLM w/o para &
Alana Everhart, known far and wide for being the record-holder for the highest-altitude skydive from the edge of space. \textit{[Correct]} \\
\addlinespace
Demasking arLLM w/ para &
The annals of time remember Alana Everhart, renowned for the record-holder for the highest-altitude skydive from the edge of space. \textit{[Correct]} \\
\bottomrule
\end{tabular}
}

\subsubsection*{Biography dataset}

\noindent\textbf{Training text:}
\begin{quote}\small
``Lois Iris Gaultney's birthday is on July 19, 1993. He began his journey in Eugene, OR. He earned his educational laurels at University of California, Berkeley. He undertook a major in Music. He devoted his time to Sherwin-Williams. He engaged in work projects in Cleveland, OH.''
\end{quote}

\noindent\textbf{Question (forward):} \textit{``What major did Lois Iris Gaultney study?''}\\
\noindent\textbf{Ground truth:} \textit{Music}

\medskip
{\small
\begin{tabular}{@{}p{0.28\linewidth}p{0.68\linewidth}@{}}
\toprule
\textbf{Method} & \textbf{Generated answer} \\
\midrule
arLLM w/o para &
Lois Iris Gaultney graduated successfully from University of California, Berkeley. She was in charge of tasks at Prudential Financial. She acknowledges her birth on June 15, 1993. She opened her eyes for the first time in Newark, NJ. She was on the job in Newark, NJ. She followed an educational track in Educational Leadership and Administration. \textit{[Wrong]} \\
\addlinespace
arLLM w/ para &
Lois Iris Gaultney studied in the field of Fine Arts and Studio Arts. He was a student at University of California, Berkeley. He was part of a program that emphasized coursework in Art and Design. He was professionally active at Sherwin-Williams. He was professionally stationed in Cleveland, OH. \textit{[Wrong]} \\
\addlinespace
dLLM w/o para &
Lois Iris Gaultney earned his educational laurels at University of California, Berkeley. He began his journey in Eugene, OR. He devoted his time to Sherwin-Williams. He undertook a major in Music. His birthday is on July 19, 1993. He engaged in work projects in Cleveland, OH. \textit{[Correct]} \\
\addlinespace
dLLM w/ para &
Lois Iris Gaultney set his academic sights on Music. He fulfilled duties at Sherwin-Williams. He received his degree from University of California, Berkeley. He was delivered in Eugene, OR. His birth is celebrated on July 19, 1993. He was active in the workforce of Cleveland, OH. \textit{[Correct]} \\
\addlinespace
Demasking arLLM w/o para &
Lois Iris Gaultney studied Music. \textit{[Correct]} \\
\addlinespace
Demasking arLLM w/ para &
Lois Iris Gaultney absorbed the subject matter of Music. He graduated from the program at University of California, Berkeley. He was delivered on July 19, 1993. He experienced professional growth in Cleveland, OH. He opened his eyes for the first time in Eugene, OR. He was a team member at Sherwin-Williams. \textit{[Correct]} \\
\bottomrule
\end{tabular}
}

\medskip
\noindent\textbf{Training text:}
\begin{quote}\small
``Curtis Chase Emley celebrates his special day on May 28, 1952. His life journey started in Elk Grove, CA. He completed his degree requirements at Kansas State University. He specialized in EMT and Paramedic. He contributed his skills to HP. He held a job in Palo Alto, CA.''
\end{quote}

\noindent\textbf{Question (backward):} \textit{``Give me the full name of the person who has the following attributes: 1) born in Elk Grove, CA, 2) majored in EMT and Paramedic, 3) worked for HP?''}\\
\noindent\textbf{Ground truth:} \textit{Curtis Chase Emley}

\medskip
{\small
\begin{tabular}{@{}p{0.28\linewidth}p{0.68\linewidth}@{}}
\toprule
\textbf{Method} & \textbf{Generated answer} \\
\midrule
arLLM w/o para &
I can provide you with information on a person who meets the specified criteria. Their full name is Sarah Brayden Baires. She was hired in Houston, TX. She first saw the light on November 2, 1998. She finished her schooling at Kansas State University. She achieved academic proficiency in EMT and Paramedic. \textit{[Wrong]} \\
\addlinespace
arLLM w/ para &
The person with the attributes you provided is Kavya Kopparapu. \textit{[Wrong]} \\
\addlinespace
dLLM w/o para &
Curtis Chase Emley held a job in Palo Alto, CA. His life journey started in Elk Grove, CA. He specialized in EMT and Paramedic. He completed his degree requirements at Kansas State University. He celebrates his special day on May 28, 1952. He contributed his skills to HP. \textit{[Correct]} \\
\addlinespace
dLLM w/ para &
Curtis Chase Emley held a job in Palo Alto, CA. He completed his degree requirements at Kansas State University. He started his life on May 28, 1952. He studied in the field of EMT and Paramedic. He contributed his skills to HP. His origins trace back to Elk Grove, CA. \textit{[Correct]} \\
\addlinespace
Demasking arLLM w/o para &
Curtis Chase Emley held a job in Palo Alto, CA. His life journey started in Elk Grove, CA. He specialized in EMT and Paramedic. He completed his degree requirements at Kansas State University. He celebrates his special day on May 28, 1952. He contributed his skills to HP. \textit{[Correct]} \\
\addlinespace
Demasking arLLM w/ para &
Curtis Chase Emley chose an academic focus in EMT and Paramedic. He attained his degree from Kansas State University. His birthday celebration is on May 28, 1952. He originated from Elk Grove, CA. He pursued his career in Palo Alto, CA. He was on the payroll of HP. \textit{[Correct]} \\
\bottomrule
\end{tabular}
}

\subsubsection*{Wiki dataset}

\noindent\textbf{Training text:}
\begin{quote}\small
``The 2025 WNBA Commissioner's Cup (known for sponsorship reasons as the 2025 WNBA Commissioner's Cup presented by Coinbase) was the WNBA's fifth Commissioner's Cup in league history. The Cup featured a transition away from the eleven-game format to a five-game format in the Eastern Conference and a six-game format in the Western Conference, with six teams in the Eastern Conference and seven in the Western Conference. The best team in cup play from each conference faced off in a championship game. The Cup was won by the Indiana Fever, the franchise's first Commissioner's Cup.''
\end{quote}

\noindent\textbf{Question (backward):} \textit{``Indiana Fever won which Cup?''}\\
\noindent\textbf{Ground truth:} \textit{2025 WNBA Commissioner's Cup}

\medskip
{\small
\begin{tabular}{@{}p{0.28\linewidth}p{0.68\linewidth}@{}}
\toprule
\textbf{Method} & \textbf{Generated answer} \\
\midrule
arLLM w/o para &
The Indiana Fever is a professional women's basketball team that plays in the Women's National Basketball Association (WNBA). They have won one WNBA championship, which was in 2012. \textit{[Wrong]} \\
\addlinespace
arLLM w/ para &
The 2025 WNBA Commissioner's Cup (recognized for sponsorship as the 2025 WNBA Commissioner's Cup presented by Coinbase) was the league's fifth Commissioner's Cup in its history. The Cup introduced a shift from the eleven-game format to a five-game series in the Eastern Conference and a six-game series in the Western Conference, featuring six teams in the Eastern Conference and seven in the Western Conference. The leading team in cup competition from each conference met in a championship game. The Cup was claimed by the Indiana Fever, the franchise's inaugural Commissioner's Cup. \textit{[Correct]} \\
\addlinespace
dLLM w/o para &
The Indiana Fever won the 2025 WNBA Commissioner's Cup. The Cup featured a transition away from the eleven-game format to a five-game format in the Eastern Conference and a six-game format in the Western Conference, with six teams in the Eastern Conference and seven in the Western Conference. The Cup was won by the Indiana Fever, the franchise's first Commissioner's Cup. \textit{[Correct]} \\
\addlinespace
dLLM w/ para &
The Indiana Fever won the Commissioner's Cup in the 2025 WNBA Commissioner's Cup. The Cup was the fifth Commissioner's Cup in the WNBA's history and introduced a shift from the eleven-game format to a five-game series in the Eastern Conference and a six-game series in the Western Conference, featuring six teams in the Eastern Conference and seven in the Western Conference. The leading team in cup competition from each conference met in a championship game. The Cup was claimed by the Indiana Fever. \textit{[Correct]} \\
\addlinespace
Demasking arLLM w/o para &
The Indiana Fever won the 2025 WNBA Commissioner's Cup. \textit{[Correct]} \\
\addlinespace
Demasking arLLM w/ para &
The 2025 WNBA Commissioner's Cup (known for sponsorship reasons as the 2025 WNBA Commissioner's Cup presented by Coinbase) was the WNBA's fifth Commissioner's Cup in league history. The Cup featured a transition away from the eleven-game format to a five-game format in the Eastern Conference and a six-game format in the Western Conference, with six teams in the Eastern Conference and seven in the Western Conference. The best team in cup play from each conference faced off in a championship game. The Cup was won by the Indiana Fever, the franchise's first Commissioner's Cup. \textit{[Correct]} \\
\bottomrule
\end{tabular}
}

\medskip
\noindent\textbf{Training text:}
\begin{quote}\small
``The 2025 Sonsio Grand Prix was the fifth round of the 2025 IndyCar season. The race was held on May 10, 2025, in Speedway, Indiana, at the Indianapolis Motor Speedway Road Course. The race was contested over 85 laps. \'Alex Palou won the race. Pato O'Ward finished 2nd, and Will Power finished 3rd. Scott McLaughlin and Scott Dixon rounded out the top five, and Graham Rahal, Marcus Armstrong, Kyle Kirkwood, Rinus VeeKay, and Felix Rosenqvist rounded out the top ten.''
\end{quote}

\noindent\textbf{Question (forward):} \textit{``How many laps was the 2025 Sonsio Grand Prix contested over?''}\\
\noindent\textbf{Ground truth:} \textit{85 laps}

\medskip
{\small
\begin{tabular}{@{}p{0.28\linewidth}p{0.68\linewidth}@{}}
\toprule
\textbf{Method} & \textbf{Generated answer} \\
\midrule
arLLM w/o para &
The 2025 Sonsio Grand Prix was contested over 90 laps. \textit{[Wrong]} \\
\addlinespace
arLLM w/ para &
The 2025 Sonsio Grand Prix was the fifth round of the 2025 IndyCar season. The race was held on May 10, 2025, in Speedway, Indiana, at the Indianapolis Motor Speedway Road Course. The race was contested over 85 laps. \'Alex Palou won the race. Pato O'Ward finished 2nd, and Will Power finished 3rd. Scott McLaughlin and Scott Dixon rounded out the top five, and Graham Rahal, Marcus Armstrong, Kyle Kirkwood, Rinus VeeKay, and Felix Rosenqvist rounded out the top ten. \textit{[Correct]} \\
\addlinespace
dLLM w/o para &
The 2025 Sonsio Grand Prix was contested over 85 laps. The race was held on May 10, 2025, in Speedway, Indiana, at the Indianapolis Motor Speedway Road Course. \'Alex Palou won the race. Pato O'Ward finished 2nd, and Will Power finished 3rd. Scott McLaughlin and Scott Dixon rounded out the top five, and Graham Rahal, Marcus Armstrong, Kyle Kirkwood, Rinus VeeKay, and Felix Rosenqvist rounded out the top ten. \textit{[Correct]} \\
\addlinespace
dLLM w/ para &
The 2025 Sonsio Grand Prix was held on May 10, 2025, in Speedway, Indiana, at the Indianapolis Motor Speedway Road Course. The race was contested over 85 laps. \'Alex Palou won the race. Pato O'Ward finished in 2nd, and Will Power finished in 3rd. Scott McLaughlin and Scott Dixon rounded out the top five. \textit{[Correct]} \\
\addlinespace
Demasking arLLM w/o para &
The 2025 Sonsio Grand Prix was contested over 85 laps. \textit{[Correct]} \\
\addlinespace
Demasking arLLM w/ para &
The 2025 Sonsio Grand Prix was the fifth round of the 2025 IndyCar season. The race was held on May 10, 2025, in Speedway, Indiana, at the Indianapolis Motor Speedway Road Course. The race was contested over 85 laps. \'Alex Palou won the race. Pato O'Ward finished 2nd, and Will Power finished 3rd. Scott McLaughlin and Scott Dixon rounded out the top five, and Graham Rahal, Marcus Armstrong, Kyle Kirkwood, Rinus VeeKay, and Felix Rosenqvist rounded out the top ten. \textit{[Correct]} \\
\bottomrule
\end{tabular}
}


\end{document}